\documentclass{article}

\usepackage{arxiv_parts/arxiv}

\usepackage{amsmath,amssymb,amsfonts}
\usepackage{algorithmic}
\usepackage{graphicx}
\usepackage{caption}
\usepackage{subcaption}
\usepackage{textcomp}
\usepackage{xcolor}
\usepackage{amsthm}
\usepackage{enumerate}
\usepackage{dsfont}
\usepackage{siunitx}
\usepackage{multirow}
\usepackage{bbm}
\usepackage{bm}
\makeatletter
\newcommand{\Spvek}[2][r]{%
  \gdef\@VORNE{1}
  \left(\hskip-\arraycolsep%
    \begin{array}{#1}\vekSp@lten{#2}\end{array}%
  \hskip-\arraycolsep\right)}

\def\vekSp@lten#1{\xvekSp@lten#1;vekL@stLine;}
\def\vekL@stLine{vekL@stLine}
\def\xvekSp@lten#1;{\def\temp{#1}%
  \ifx\temp\vekL@stLine
  \else
    \ifnum\@VORNE=1\gdef\@VORNE{0}
    \else\@arraycr\fi%
    #1%
    \expandafter\xvekSp@lten
  \fi}
\makeatother

\newtheorem{Theorem}{Theorem}
\newtheorem{Proposition}{Proposition}

\newcommand{\W}{\mathcal{W}}
\newcommand{\Deellip}{\textit{DEEL.LIP}\footnote{https://github.com/deel-ai/deel-lip}}

\usepackage[pagebackref=true,breaklinks=true,colorlinks,bookmarks=false]{hyperref}

\begin{document}

\title{Achieving robustness in classification using optimal transport with hinge regularization}
\author{%
  Mathieu Serrurier\\
  IRIT\\
  Université Paul Sabatier Toulouse
  \And
  Franck Mamalet\\
  IRT Saint-Exupery
  \And
  Alberto González-Sanz\\
  IMT\\
  Université Paul Sabatier Toulouse
  \And
  Thibaut Boissin\\
  IRT Saint-Exupery
  \And
  Jean-Michel Loubes \\
  IMT\\
  Université Paul Sabatier Toulouse
  \And
  Eustasio del Barrio \\
  Dpto. de Estadística e Investigación Operativa\\
  Universidad de Valladolid
  
}
\maketitle
\begin{abstract}

Adversarial examples have pointed out Deep Neural Networks vulnerability to small local noise. It has been shown that constraining their Lipschitz constant should enhance robustness, but make them harder to learn with classical loss functions. We propose a new framework for binary classification, based on optimal transport, which integrates this Lipschitz constraint as a theoretical requirement. We propose to learn 1-Lipschitz networks using a new loss that is an hinge regularized version of the Kantorovich-Rubinstein dual formulation for the Wasserstein distance estimation. This loss function has a direct interpretation in terms of adversarial robustness together with certifiable robustness bound. We also prove that this hinge regularized version is still the dual formulation of an optimal transportation problem, and has a solution. We also establish several geometrical properties of this optimal solution, and extend the approach to multi-class problems. Experiments show that the proposed approach provides the expected guarantees in terms of robustness without any significant accuracy drop. The adversarial examples, on the proposed models, visibly and meaningfully change the input providing an explanation for the classification.


\end{abstract}

\section{Introduction}
\label{sec:intro}
The important progress in deep learning has led to a massive interest for these approaches in industry. However, when applying machine learning to critical tasks such has in the transportation or the medical domain, empirical and theoretical guarantees are required. Unfortunately, it has been shown that neural networks are weak to adversarial attacks: a carefully chosen small shift to the input, usually indistinguishable from noise, can change the class prediction \cite{szegedy2013}.  This sensitivity to adversarial attacks is mainly due to the Lipschitz constant of a neural network which can be arbitrarily high when unconstrained. Most of white-box attacks (where the full model is available) take advantage of it to build adversarial examples by using gradient descent with respect to the input variables. $FGSM$~\cite{Goodfellow2014a} performs only one step of gradient descent when other approaches such as $PGD$~\cite{Madry2017,carlini2017towards} find the optimal adversarial example iteratively. In black-box scenarios, gradients or logits of the model are not available. In such case, attacks start from large perturbations and then reducing it step by step (see for instance, boundary attacks~\cite{Brendel2018} and pointwise attacks~\cite{Schott2018}).

There are three major types of strategy to address the issue of adversarial attacks. Agnostic defenses are independent of the model and consist in altering the input or the prediction. For instance, Cohen et al. obtain a provable certificate with respect to $l_2$ norm by using Gaussian random smoothing~\cite{pmlr-v97-cohen19c}. DEFENSE-GAN~\cite{samangouei2018defensegan} uses a GAN to transform the input into the closest non-adversarial one at inference time. The second group of strategies relies on saddle point optimization by adding a penalty term measuring the empirical weakness against adversarial example during the learning process~\cite{Madry2017}. The last type of approaches focus on the Lipschitz constant of the network. It has been proven that bounding the  Lipschitz constant of a neural network provides certifiable robustness guarantees against local adversarial attacks~\cite{hein_formal_2017,ono_lightweight_2018}, improves generalizations~\cite{Sokolic_2017} and the interpretability of the model~\cite{tsipras2018robustness}.  This constraint can be achieved layer by layer by using spectral normalization~\cite{cisse_parseval_2017} or non-expansive layer~\cite{qian_l2-nonexpansive_2019}. In \cite{li2019preventing}, Li et al. go beyond the Lipschitz constant bounding by requiring layers to be gradient norm preserving. Combined with hinge loss, it allows them to achieve stronger robustness certification. However, the main limitation of this approach relies in the link between the hinge margin parameter and the robustness of the network.

In this paper we propose a new classification framework based on optimal transport that integrates the Lipschitz constant and the gradient norm preserving constraint as a theoretical requirement. To the best of our knowledge, very few researches investigate the link between binary classification and optimal transport (in~\cite{Frogner2015}, Frogner et al. use Wasserstein loss to improve multilabel classification). In Wasserstein GAN~\cite{Arjovsky2017}, the k-Lipschitz networks used to measure the distance between two distributions act like a discriminator, in analogy with the initial GAN algorithm \cite{Goodfellow2014}. The Wasserstein distance is approximated using a loss based on the Kantorovich-Rubinstein dual formulation and a k-Lipschitz network constrained by weight clipping. However, as we will demonstrate, a vanilla classifier based on the Kantorovich-Rubinstein loss is suboptimal, even on toy datasets.

We propose a binary classifier based on a regularized version of Kantorovich-Rubinstein formulation using a hinge loss term. We show that it remains the dual of an optimal transport problem, and we prove that the optimal solution of the problem exists and makes no error when the classes are well separated. With this new optimization problem, we guarantee to have an accurate classifier with a loss that is defined on and takes advantage of 1-Lipschitz function. As in~\cite{li2019preventing}, we bound the Lipschitz constant of the linear layers by B\"jorck normalization and use norm preserving activation functions~\cite{pmlr-v97-anil19a}. However, the optimal transport interpretation of the problem makes the bridge between these constraints and the loss function. When solving this optimal transport problem, attacking a prediction corresponds to travel along the transport plan up to the decision frontier. The output of the optimal network is linked to the length of this path, which is maximized during the optimization process of the proposed loss. 

The paper, and the contributions, are structured as follows. In Section~\ref{sec:wass_distance}, we recall the definition of Wasserstein distance and the dual optimization problem associated. We present the interesting properties of a classifier based on this approach, illustrate that it leads to a suboptimal classifier even on a toy dataset. Section~\ref{sec:kr_hinge} describes the proposed binary classifier, based on a regularized version of the Kantorovich-Rubinstein loss with a hinge regularization. We show that the primal of this classification problem is a new optimal transport problem and we demonstrate different mathematical properties of our approach. Section~\ref{sec:lip_const} is devoted to the way of constraining the classifier to be  1-Lipschitz and how to generalize the approach to multi classification problems. Section~\ref{sec:experimentation} presents the results of  experiments on MNIST, Cifar10 and CelebA datasets, measuring and comparing the results of different approaches in terms of accuracy and robustness. Last, we demonstrate that with our approach, building an adversarial example requires explicitly changing the example to an in-between two-classes image, which correspond to a point halfway on the transport plan. Proofs, computations details and additional experiments are reported in the appendix.

\section{Wasserstein distance and Kantorovich-Rubinstein classifier}
\label{sec:wass_distance}
In this paper we only consider the Wasserstein-1 distance, also called Earth-mover, and noted $\W$ for $\W_1$.
The $1$-Wasserstein distance between two probability distributions $\mu$ and $\nu$ in $\Omega$, and its dual formulation by Kantorovich-Rubinstein duality \cite{villani2008}, is defined as the solution of:
\begin{subequations}

\begin{align}
\W(\mu,\nu) & = \inf_{\pi \in \Pi(\mu,\nu)}\underset{x,z \sim \pi}{\mathbb{E}}\parallel \textbf{x}-\textbf{z} \parallel \label{wasserstein}\\
  & =\sup_{f \in Lip_1(\Omega)} \underset{\textbf{x} \sim \mu}{\mathbb{E}} \left[f(\textbf{x} )\right] -\underset{\textbf{x}  \sim \nu}{\mathbb{E}} \left[f(\textbf{x} )\right] \label{kantorovich}
\end{align}
\end{subequations}

where $\Pi(\mu,\nu)$ is the set of all probability measures on $\Omega\times \Omega$ with marginals $\mu$ and $\nu$ and $Lip_1(\Omega)$ denotes the space of 1-Lipschitz functions over $\Omega$. 
Although, the infimum in Eq.~\eqref{wasserstein} is not tractable in general, the dual problem can be estimated through the optimization of a regularized neural network. This approach has been introduced in WGAN  \cite{Arjovsky2017} where $Lip_1(\Omega)$ is approximated by the set of neural networks with bounded weights (better approximations of $Lip_1(\Omega)$ will be discussed in Section~\ref{sec:lip_const}). 

We consider a binary classification problem on feature vector space $X\subset \Omega$ and and labels $Y= \{-1,1\}$. We name $P_+=\mathbb{P}(X|Y=1)$ and  $P_-=\mathbb{P}(X|Y=-1)$, the conditional distributions with respect to Y. We note $p=P(Y=1)$ and $1-p=P(Y=-1)$ the apriori class distribution. The classification problem is balanced when $p=\frac{1}{2}$.\\

In WGAN, \cite{Arjovsky2017} proposed to use the learned neural network (denoted $\hat{f}$ in the following),  by maximizing the Eq.~\eqref{kantorovich}, as a discriminator between fake and real images, in analogy with GAN~\cite{Goodfellow2014}.
To build a classifier based on $\hat{f}$, one can simply note that if  $f^*$ is an optimal solution of Eq.~\eqref{kantorovich}, then  $f^* + C, C\in \mathbb{R}$, is also optimal. Centering the function $f^*$ (resp. $\hat{f}$), Eq.~\eqref{eq:wass_vanilla_classif}, enables classification according to the sign of $f_c^*(x)$ (resp.$\hat{f_c}$ for the empirical solution).
\begin{equation}
\label{eq:wass_vanilla_classif}
f_c^*(\textbf{x} )=f^*(\textbf{x} )-\frac{1}{2}\left(\underset{\textbf{z}  \sim P_+}{\mathbb{E}} \left[f^*(\textbf{z} )\right] +\underset{\textbf{z}  \sim P_-}{\mathbb{E}} \left[f^*(\textbf{z} )\right]\right).
\end{equation}

Such a classifier would exhibit good properties in terms of robustness for two main reasons: First, it has been shown in \cite{villani2008} that the function $f^*$ is directly related to the cost of transportation between two points linked by the transportation plan 
as follows:


\begin{equation}
\label{eq:transport_cost}
\mathbb{P}_{\textbf{x},\textbf{z}\sim \pi^*}(f^*(\textbf{x})-f^*(\textbf{z})=||\textbf{x}-\textbf{z}||)=1.
\end{equation}
 Second, it was shown in \cite{gulrajani2017improved,pmlr-v97-anil19a}, that this optimal solution also induces a property stronger  than 1-Lipschitz:
\begin{equation}
\label{eq:nabla_1}
||\nabla f^* ||=1 \text{ almost surely on the support of }\pi^*.
\end{equation}

\begin{figure*}
\centering
\begin{subfigure}{.5\textwidth}
  \centering
  \includegraphics[width=1\linewidth]{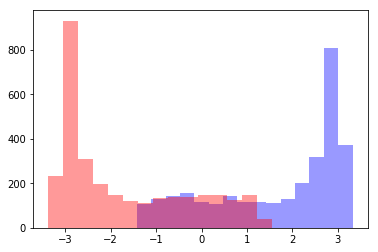}
  \caption{Distribution of $\hat{f_c}$ conditionally to the classes}
  \label{wass:sub1}
\end{subfigure}%
\begin{subfigure}{.5\textwidth}
  \centering
  \includegraphics[width=0.9\linewidth]{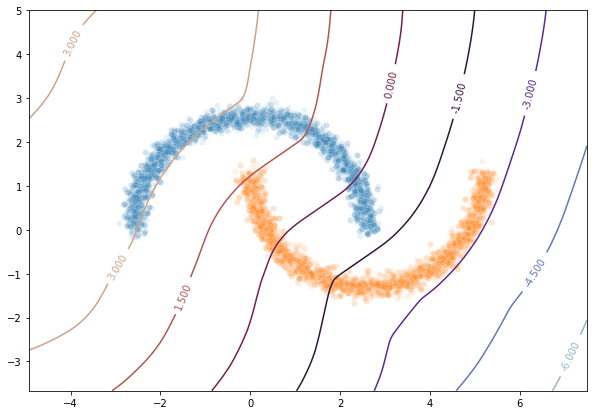}
  \caption{Level map of $\hat{f_c}$}
  \label{wass:sub2}
\end{subfigure}
\caption{Wasserstein classification (Eq.~\eqref{eq:wass_vanilla_classif}) on the two moons.}
\label{fig:wass}
\end{figure*}

However, applying this vanilla classifier (Eq.~\eqref{eq:wass_vanilla_classif}) to a toy dataset such as the two-moons problem, leads to a poor accuracy.
Indeed, Figures~\ref{wass:sub1} and~\ref{wass:sub2} present respectively the distribution of the values of $\hat{f_c}(x)$ conditionally to the classes and the level map of $\hat{f_c}$. We can observe that, even if the classes are easily separable, the distributions of the values of $\hat{f_c}$ conditionally to the class overlap. Thus, the 0-level threshold on $\hat{f_c}$ does not correspond to the optimal separator (even if it is better than random). Intuitively, $\hat{f_c}$  
maximizes the difference of the expectancy of the image of  
the two distributions but do not try to minimize their overlap (Fig.~\ref{wass:sub1}). 

\section{Hinge regularized Kantorovich-Rubinstein classifier }
\label{sec:kr_hinge}
\subsection{Definitions and primal transportation problem}
In order to improve the classification abilities of the classifier based on Wasserstein distance, we propose a Kantorovich-Rubinstein optimization problem regularized by an hinge loss :

\begin{equation} \label{eq:reg_OT} 
\sup_{f \in Lip_1(\Omega)} -\mathcal{L}^{hKR}_\lambda(f) = \inf_{f \in Lip_1(\Omega)} \underset{\textbf{x} \sim P_-}{\mathbb{E}} \left[f(\textbf{x})\right]-\underset{\textbf{x} \sim P_+}{\mathbb{E}}\left[f(\textbf{x})\right] +\lambda\underset{\textbf{x}}{\mathbb{E}} \left(1-Yf(\textbf{x})\right)_+ 
\end{equation}

where $(1-\textbf{y}f(\textbf{x}))_+$ stands for the hinge loss $max(0,1-\textbf{y}f(\textbf{x}))$ and $\lambda>=0$. We name $\mathcal{L}^{hKR}_\lambda$ the hinge-KR loss. The goal is then to minimize this loss with an 1-Lipschitz neural network.
When $\lambda=0$, this corresponds to the Kantorovich-Rubinstein dual optimization problem. Intuitively, the 1-Lipschitz function $f^*$ optimal with respect to Eq.~\eqref{eq:reg_OT} is the one that both separates the examples with a margin and spreads as much as possible the image of the distributions. When using only an hinge loss (as in \cite{li2019preventing} for instance), the examples outside the margin are no more taken into consideration. If the margin is increased to cover all the examples and if the class are equally distributed, the hinge loss becomes equivalent to the Kantorovich Rubinstein loss and then leads to a weak classifier. 

In the following, we introduce Theorems that prove the existence of such an optimal function $f^*$ and important properties of this function. Demonstrations of these theorems are in Appendix~\ref{appendix:proofs}.
\begin{Theorem} [Solution existence]
\label{Lemma:bounded_M}
 For each $\lambda>0$ there exists at least a solution $f^*$ to the problem  \[ f^*:=f_{\lambda}^*\in{\arg\min}_{f\in \text{Lip}_1(\Omega)}\mathcal{L}^{hKR}_{\lambda}(f) . \]
\end{Theorem}
Moreover, let $\psi$ be an optimal transport potential for the transport problem from $P_+$ to $ P_-$, $f^*$ satisfies that 
\begin{align}\label{eq:M}
	|| f^*||_{\infty}\leq M:= 1+\text{diam}(\Omega)+\frac{L_1(\psi)}{\inf(p,1-p)}.
\end{align}
The next theorem establishes that the Kantorovich-Rubinstein optimization problem with hinge regularization is still a transportation problem with relaxed constraints on the joint measure (which is no longer a joint probability measure).
\begin{Theorem}[Duality]\label{Theo:dual}
Set  $P_+, P_-\in \mathcal{P}(\Omega)$ and $\lambda>0$, then the following equality holds
\begin{align}\label{eq:dual}
\begin{split}
 \sup_{f\in \text{Lip}_1(\Omega)}- \mathcal{L}^{hKR}_\lambda(f)=&	\inf_{\pi \in\Pi^p_{\lambda}(P_+, P_-)}\int_{\Omega\times \Omega}|{\textbf{x}}-{{\textbf{z}}} |d\pi + \pi_{\textbf{x}}(\Omega)+\pi_{{\textbf{z}}}(\Omega)-1
	\end{split}
\end{align}
Where $\Pi^p_{\lambda}(P_+, P_-)$ is the set consisting of positive measures $\pi\in \mathcal{M}_+(\Omega\times \Omega)$ which  are absolutely continuous with respect to the joint measure $dP_+\times dP_-$ and $\frac{d\pi_{\textbf{x}}}{dP_+}\in [p, p(1+\lambda)]$, $\frac{d\pi_{{\textbf{z}}}}{dP_-}\in [1-p, (1-p)(1+\lambda)] $.
\end{Theorem}

\subsection{Classification and geometrical properties}
\label{sec:khr_properties}

\begin{figure*}
\centering
\begin{subfigure}{.5\textwidth}
  \centering
  \includegraphics[width=1\linewidth]{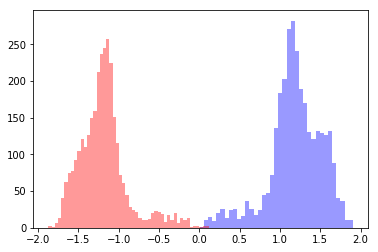}
  \caption{Distribution of $\hat{f}$ conditionally to the classes.}
  \label{wass_pen:sub1}
\end{subfigure}%
\begin{subfigure}{.5\textwidth}
  \centering
  \includegraphics[width=0.9\linewidth]{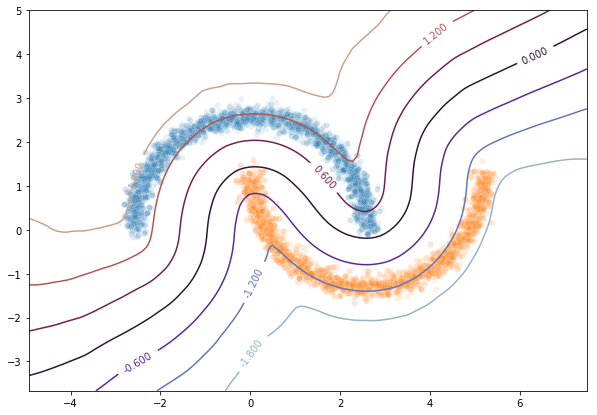}
  \caption{Level map of $\hat{f}$}
  \label{wass_pen:sub2}
\end{subfigure}
\caption{Hinge regularized Kantorovich-Rubinstein (hinge-KR) classification on the two moons problem}
\label{fig:wass_pen}
\end{figure*}
We note $\hat{f}$ the solution obtained by minimizing $\mathcal{L}^{hKR}_\lambda$ on a set of labeled examples and $f^*$ the solution of Eq.~\eqref{eq:reg_OT}. We don't assume that the solution found is optimal (i.e. $\hat{f}\neq f^*$) but we assume that $\hat{f}$ is 1-Lipschitz. Given a function $f$, a classifier based on $sign(f)$ and an example $x$, an adversarial example is defined as follows:
\begin{equation}
    adv(f,\textbf{x})=\underset{\textbf{z}\in \Omega | sign(f(\textbf{z}))=-sign(f(\textbf{x}))}{argmin}\parallel \textbf{x}-\textbf{z}\parallel.
\end{equation}
According to the 1-Lipschitz property of $\hat{f}$ we have 
\begin{equation}
\label{lowerbound}
|\hat{f}(\textbf{x})| \leq |\hat{f}(\textbf{x})-\hat{f}(adv(\hat{f},\textbf{x}))| \leq \parallel \textbf{x}- adv(\hat{f},\textbf{x}) \parallel.
\end{equation}
So $|\hat{f}(\textbf{x})|$ is a lower bound of the distance of $x$ to the separating boundary defined by $\hat{f}$ and thus a lower bound to the robustness to $l_2$ adversarial attacks. Thus, by minimizing $\mathbb{E} \left((1-\textbf{y}f(\textbf{x}))_+\right)$, we maximize the accuracy of the classifier and by maximizing the discrepancy of the image of $P_+$ and $P_-$ with respect to $f$ we maximize the robustness with respect to adversarial attack. The proposition below establishes that the gradient of the optimal function with respect to Eq.~\eqref{eq:reg_OT} has norm 1 almost surely, as for the unregularized case (Eq.~\eqref{eq:nabla_1}).
\begin{Proposition}\label{coro_norm1}
Let $\pi$ be the optimal measure of the dual version \eqref{eq:dual} of the hinge regularized optimal transport problem. Suppose that it is absolutely continuous with respect to Lebesgue measure. Then there exists at least a solution $f^*$ of \eqref{eq:dual} such that $||\nabla f^*||=1$ almost surely.
\end{Proposition}
Furthermore, empirical results suggest that given $\textbf{x}$, the image $tr_{f^*}(\textbf{x})$ of $\textbf{x}$ by transport plan and $adv(\textbf{x})$ are in the same direction with respect to $\textbf{x}$ and the direction is 
 $-\nabla_x f^*(\textbf{x})$. Combining this direction with the Eq.~\eqref{lowerbound}, we will show empirically (sect.~\ref{sec:experimentation}) that  $$adv(\textbf{x})\approx x-c_x*f^*(\textbf{x})*\nabla_x f^*(\textbf{x})$$ and $$tr_{f^*}(\textbf{x})\approx x-c'_x*f^*(\textbf{x})*\nabla_x f^*(\textbf{x})$$ with $1 \leq c_x \leq c_x' \in \mathbb{R} $. It turns out that this corresponds to the definition of FGSM attacks \cite{Goodfellow2014a}. This suggests that in our frameworks, adversarial attacks amount to travel along the transportation path from the example to its transportation image.
 
The next proposition shows that, if the classes are well separated, maximizing the hinge-KR loss leads to a perfect classifier.
\begin{Proposition}[Separable classes]\label{coro:epsilon}
Set  $P_+, P_-\in \mathcal{P}(\Omega)$ such that $P(Y=+1)=P(Y=-1)$ and $\lambda\geq 0$ and suppose that there exists $\epsilon>0$ such that
\begin{align}\label{coro:hipotesis}
	| {\textbf{x}}-{\textbf{{{\textbf{z}}}}} |> 2 \epsilon \ \ dP_+\times dP_- \text{ almost surely}
\end{align}
Then for each
$$
	f_{\lambda}\in {\arg\sup}_{f\in \text{Lip}_{1/\epsilon}(\Omega)}\int_{\Omega}f(dP_+-dP_-)-
	\lambda \left( \int_{\Omega}(1-f)_+dP_+ +\int_{\Omega}(1+f)_+dP_-\right),
$$
it is satisfied that $L_1(f_{\lambda})=0.$ Furthermore if $\epsilon\geq 1$ then $f_{\lambda}$ is an optimal transport potential from $P_+$ to $P_-$ for the cost $| {\textbf{x}}-{{\textbf{z}}}|$.
\end{Proposition}
We show in Fig.~\ref{fig:wass_pen}, on the two moons problem, that in contrast to the vanilla classifier based on Wasserstein (Eq.~\eqref{eq:wass_vanilla_classif}), the proposed approach enables non overlapping distributions of $\hat{f}$ conditionally to the classes (Fig.~\ref{wass_pen:sub1}).
In the same way, the 0-level cut of $\hat{f}$ (Fig.~\ref{wass_pen:sub2}) is a nearly optimal classification boundary. Moreover, the level cut of $\hat{f}$, on the support of the distributions, is close to the distance to this classification boundary.

\section{Architecture}
\label{sec:lip_const}
\subsection{1-Lipschitz gradient norm preserving network}
In order to build a deep learning classifier based on the hinge-KR optimization problem (Eq.~\eqref{eq:reg_OT}), we have to constrain the Lipschitz constant of the neural network to be equal to 1.  It is known that evaluating it exactly is a NP-hard problem~\cite{NIPS2018_7640}. The simplest way to constraint a network to be 1-Lipschitz is to impose this 1-Lipschitz property to each layer. For dense layers, the initial version of WGAN \cite{Arjovsky2017} consisted of clipping the weights of the layers. However, this is a very crude way to upper-bound Lipschitz constant. Normalizing by the Frobenius norm has also been proposed in~\cite{SalimansK16}. In this paper, we use spectral normalization as proposed in~\cite{Miyato2018SpectralNF}, since the spectral norm is equals to the Lipschitz constant of the layer. At the inference step, we normalize the weights of each layer by the spectral norm of the matrix. This spectral norm is computed by iteratively evaluating the largest singular value with the power iteration algorithm~\cite{GOLUB200035}. This is done during the forward step and taken into account for the gradient computation. In the case of 2D-convolutional layers, normalizing by the spectral norm of convolution kernels is not enough and a supplementary multiplicative constant $\Lambda$ is required (the regularization is then done by dividing $W$ by $\Lambda||W||$). We propose, for zero padding, a tighter estimation of $\Lambda$ than the one proposed in~\cite{cisse_parseval_2017}, computing the average duplication factor of non zero padded values in the feature map:

\begin{equation}
\Lambda=\sqrt{\frac{(k.w-\bar{k}.(\bar{k}+1)).(k.h-\bar{k}.(\bar{k}+1))}{h.w}}
\label{eq:ConvApproxLip}
\end{equation}
for a kernel size equals to $k=2*\bar{k}+1$.
Even if this constant doesn't provide a strict upper bound of the Lipschitz constant (for instance, when the higher values are located in the center of the picture), it behaves very well empirically.
Convolution with stride, pooling layers, detailed explanations and demonstrations are discussed in Appendix~\ref{sec:convStride}.\\

As shown in Property~\ref{coro_norm1}, the optimal function $f^*$ with respect to Eq.~\eqref{eq:reg_OT}, verifies $||\nabla f^* ||=1$ almost surely (\textit{gradient norm preserving} (GNP) architecture~\cite{pmlr-v97-anil19a} ). We apply the approach described in~\cite{pmlr-v97-anil19a}, based on the use of specific activation functions and a process of normalization  of the weights. Two norm preserving activation functions are proposed: i)\ \textbf{GroupSort2} : sorting the vector by pairs, ii) \textbf{FullSort} : sorting the full vector.
These functions are vector-wise rather than element-wise. We also use the P-norm pooling~\cite{boureau2010}, with $P=2$ which is a norm-preserving average pooling.
Concerning linear functions, a weight matrix $W$ is gradient norm preserving if and only if all the singular values of $W$ are equals to $1$. In \cite{pmlr-v97-anil19a}, the authors propose to use the Björck orthonormalization algorithm \cite{bjorck71Ortho}. This algorithm is fully differential and, as for spectral normalization, is applied during the forward inference, and taken into account for back-propagation (see Appendix~\ref{app:normpreserving} for details). We don't consider BCOP \cite{qian_l2-nonexpansive_2019}, which performs slightly better that Björck for convolution but at an higher computation cost. We developed a full tensorflow~\cite{tensorflow2015-whitepaper} implementation in an opensource library, called \Deellip, that enables training of $k$-Lipschitz neural networks, and exporting them as standard layers for inference. 
\subsection{Multi-class hinge-KR classifier}
To adapt our approach to the multi-class case, we propose to use $q$ binary one-vs-all classifiers, where $q$ is the number of classes. The set of labels is now $Y=\{C_1,\ldots,C_q\}$. We name $P_k=\mathbb{P}(X|Y= C_k)$ and $\neg P_k=\mathbb{P}(X|Y\not = C_k)$ the  conditional  distributions  with  respect  to  $Y$. We obtain the following optimization problem :
\begin{equation}
\label{eq:hkr_multi}
\begin{split}
&\underset{f_1,\ldots,f_q \in Lip_1(\Omega)}{\text{sup}}-\mathcal{L}^{hKR}_\lambda(f_1,\ldots,f_q) \\
&\text{where}\\
&\mathcal{L}^{hKR}_\lambda(f_1,\ldots,f_q) = \sum_{k=1}^q \left[\underset{\textbf{x}  \sim \neg P_k}{\mathbb{E}} \left[f_k(\textbf{x})\right] -\underset{\textbf{x} \sim P_k}{\mathbb{E}}\left[f_k(\textbf{x})\right]   +\lambda\underset{\textbf{x}}{\mathbb{E}} \left(m-(2*\underset{y=C_k}{\mathbbm{1}} -1).f_k(x)  \right)_+ \right].
\end{split}
\end{equation}
The global architecture is the same as the binary one except that the last layer has $q$ outputs. For this last layer, each weights corresponding to each output neuron are normalized independently creating $q$ 1-Lipschitz functions with gradient norm preserving properties. With this architecture, the optimal transport interpretation is still valid. The class predicted corresponds to the argmax of the classifier functions. With this approach, the provable robustness lower-bound is the half of the difference between the max and the second max values of $\{f_1(x),\ldots,f_q(x)\}$.  

In \cite{li2019preventing}, Li et al. use classical multi-class hinge loss based on un-centered margin and apply constraint on the entire last layers rather than doing it independently. It allows to a lower provable adversarial robustness bound than ours. However, the $f_1,\ldots,f_q$ obtain in this setting are no more 1-Lipschitz one by one, making the adversarial robustness bounds not comparable directly (see Appendix~\ref{sec:robustness_bounds}).

\section{Experiments}
\label{sec:experimentation}
\begin{figure*}
\centering
\begin{subfigure}{.24\textwidth}
  \centering
  \includegraphics[width=1\linewidth]{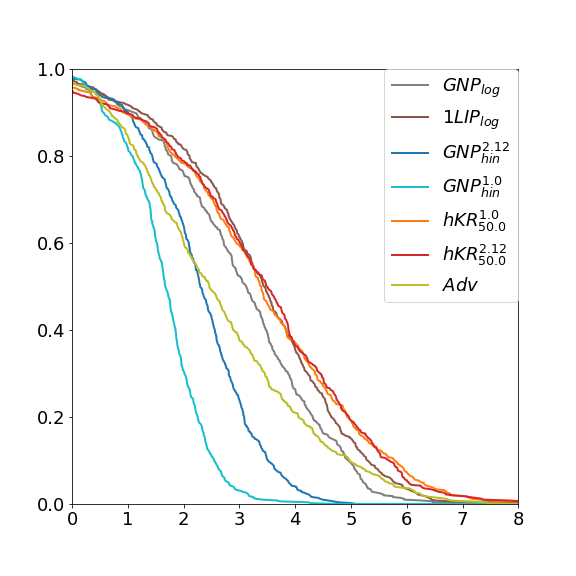}
  \caption{MNIST dense}
  \label{fig:l2norm_MLP_wass}
\end{subfigure}%
\begin{subfigure}{.24\textwidth}
  \centering
  \includegraphics[width=1\linewidth]{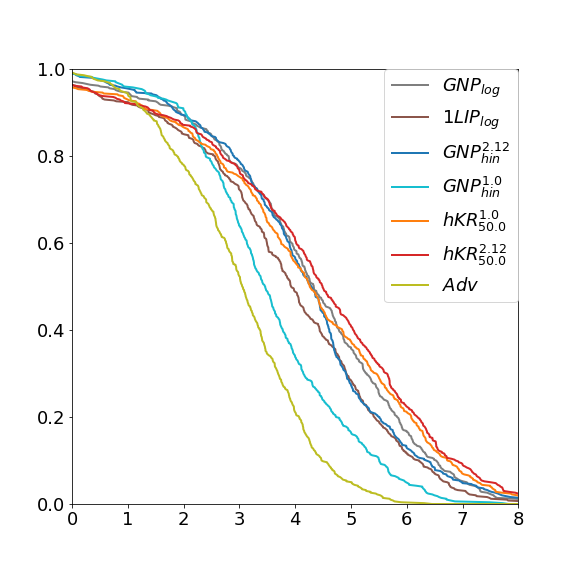}
  \caption{MNIST CNN}
  \label{fig:l2_norm_VGG_wass}
\end{subfigure}
\begin{subfigure}{.24\textwidth}
  \centering
  \includegraphics[width=1\linewidth]{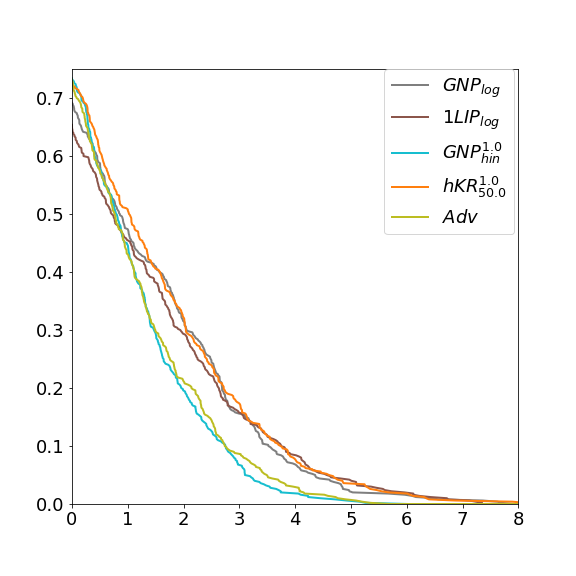}
  \caption{CIFAR}
  \label{fig:l2_norm_MLP_lipsigmoid}
\end{subfigure}
\begin{subfigure}{.24\textwidth}
  \centering
  \includegraphics[width=1\linewidth]{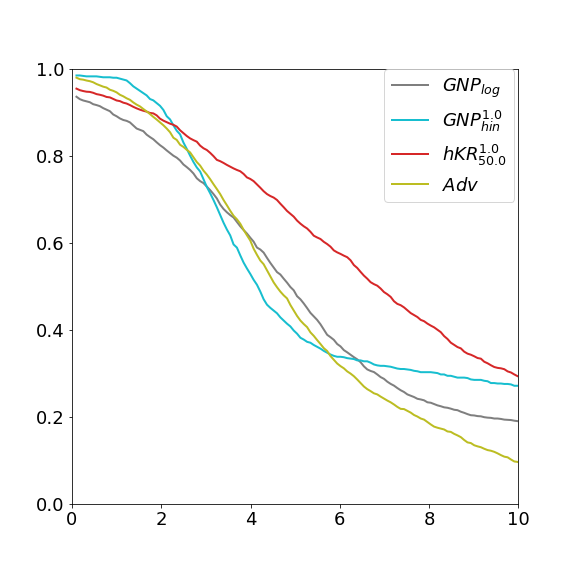}
  \caption{CelebA}
  \label{fig:l2_norm_MLP_lipsigmoid}
\end{subfigure}
\caption{Accuracy (Y-axis) w.r.t. of $l_2$ norm of fooling noise with deepfool attack (X-axis) on 2000 images of the test set
}
\label{fig:deepfool_curves}
\end{figure*}

In the experiment, we compare five approaches: i) $Adv$ for Adversarial learning as in~\cite{Madry2017} , ii) $1LIP_{log}$ for log-entropy classifier with Björck orthonormalization and ReLU activation functions similar to Parseval networks~\cite{cisse_parseval_2017}, iii) $GNP^m_{hin}$ for gradient norm preserving classifiers based on hinge loss with margin $m$ as done in~\cite{li2019preventing} iv) $GNP_{log}$ for gradient norm preserving classifiers based on log entropy loss and v) $hKR^m_\alpha$ for gradient norm preserving classifiers based on the proposed hinge-KR with margin $m$ and coefficient $\alpha$. Note that, to the best of our knowledge, the $GNP_{log}$ hasn't been applied yet for adversarial defenses. To have a fair comparison, all the classifiers share the same dense or convolutional architectures except for the weight normalization and the activation functions. We set $\alpha$ to $50$ and margin $m$ to $1$ except for MNIST where we also consider $m=2.12$ to be comparable with the experiments in~\cite{li2019preventing}. For the $GNP$ classifiers, we apply Björck orthonormalization (15 steps with p=1) after the spectral normalization (this improves the convergence of the Björck algorithm). We use fullsort activation function for dense layers, and GroupSort2 for the convolutional ones.  Appendix~\ref{sec:networks_architecture} provides the full description of the architecture and the optimization process.

We consider three classification problems, two multi-class problems (MNIST~\cite{lecun-mnisthandwrittendigit-2010} and CIFAR-10~\cite{Krizhevsky09learningmultiple}) and a binary one (eyeglasses detection in Celeba-A dataset \cite{liu2015faceattributes}). For MNIST and CIFAR-10, we use standard configurations with 10 classes and no data augmentation, $50 000$ examples in the training set and $10 000$ examples in the test set. In the binary problem, we consider the separation between people with or without eyeglasses on the CelebA dataset with 128x128x3 centered images. This is an unbalanced classification problem with $16914$ examples in the training set ($38\%$ with eyeglasses) and $16914$ examples in the test set ($38\%$ with eyeglasses).

\begin{figure*}
\centering
\begin{subfigure}{.24\textwidth}
  \centering
  \includegraphics[width=1\linewidth]{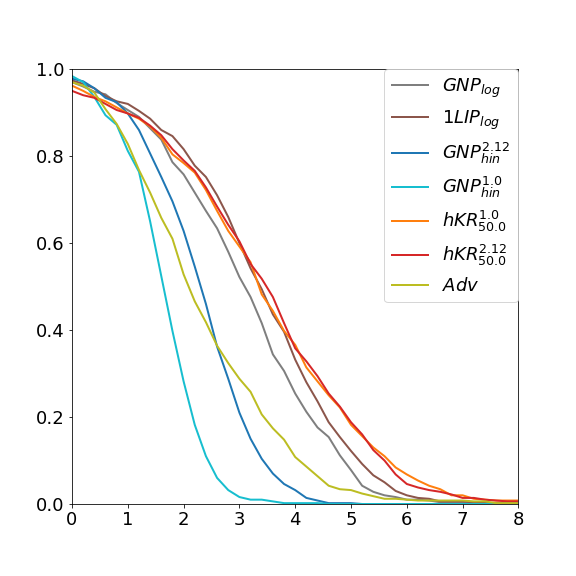}
  \caption{MNIST dense}
\end{subfigure}%
\begin{subfigure}{.24\textwidth}
  \centering
  \includegraphics[width=1\linewidth]{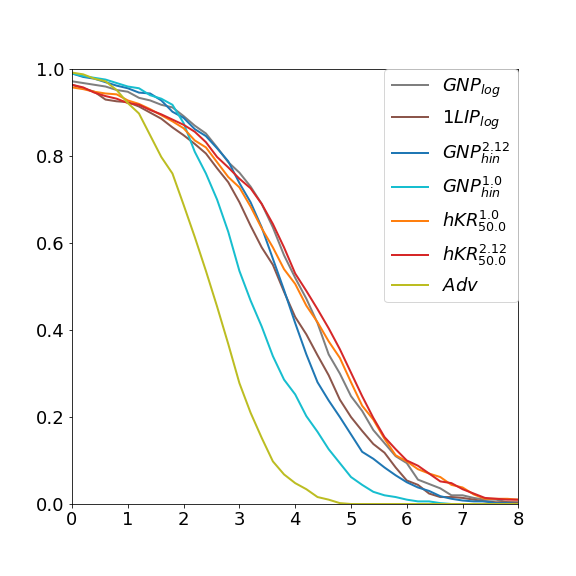}
  \caption{MNIST CNN}
\end{subfigure}
\begin{subfigure}{.24\textwidth}
  \centering
  \includegraphics[width=1\linewidth]{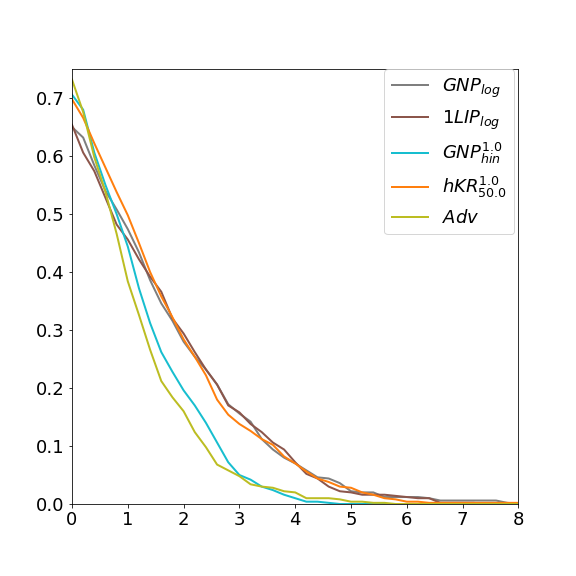}
  \caption{CIFAR}
\end{subfigure}
\begin{subfigure}{.24\textwidth}
  \centering
  \includegraphics[width=1\linewidth]{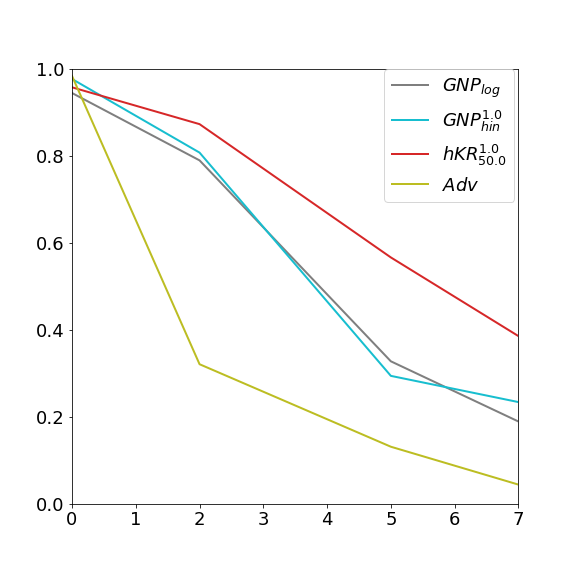}
  \caption{CelebA}
\end{subfigure}
\caption{Accuracy (Y-axis) w.r.t. of $l_2$ norm of $FGSM$, $l_2PGD$, deepfool and $l_2$ Carlini and Wagner combined attacks on 500 images of the test set
}
\label{fig:combine_curves}
\end{figure*}

Figure \ref{fig:deepfool_curves} presents the robustness against $l_2$ attacks with \textit{DeepFool} attack~\cite{moosavi-deepfool15} on the different datasets. We use this type of attack since none of the tested approaches is specifically built to resist to it. We can observe that the $hKR^m_\alpha$ approach is at the top of robustness on all the dataset and systematically better than $GNP_{log}$ on all datasets and $GNP^m_{hin}$ on the multiclass problem for the same value of $m$. On the CelebA dataset, $GNP^m_{hin}$ and the adversarial approach $Adv$ start with a better accuracy, but don't resist to large attacks. $GNP_{log}$ and $1LIP$ have good performances on MNIST and CIFAR but their performances decrease when the models are deeper as for CelebA dataset ($1LIP$ was unable to converge on CelebA). To compare the different defense methods, we also apply a combination of the state-of-the art attacks $FGSM$~\cite{Goodfellow2014a}, $PGD$~($l_2PGD$) \cite{Madry2017} and  Carlini and Wagner ($l_2CW$) \cite{carlini2017towards} on 500 images of the test set. All attacks are performed with the  foolbox library~\cite{rauber2017foolbox}. For each value of $\epsilon$, we run all the attacks and consider it as a success if at least one of them has succeeded. The results are presented in Figure~\ref{fig:combine_curves} and Table~\ref{tab:celeba_comparison} details the robustness values for CelebA. The results confirm and amplify the ones obtained with deepfool attacks. $hKR^m_\alpha$ obtain the highest level of robustness in all the situations for low and high values of $\epsilon$. 
$Adv$ has the highest robustness w.r.t. $FGSM$ attacks, since it was designed against to, but is a bit less resistant w.r.t. $l_2PGD$ and weak w.r.t. $l_2CW$. $hKR^m_\alpha$ is especially strong w.r.t. Carlini and Wagner attack even with high values of $\epsilon$. The discrepancy between the robustness of the approaches increase when models become deeper. On the CelebA dataset,  $GNP_{log}$ have difficulties to resist to higher noise. As expected, $hKR^m_\alpha$ seems to take most advantage of the gradient norm preserving architecture and obtains robustness against large attacks with an acceptable decrease of accuracy without noise. 

\begin{figure*}
\centering

    \includegraphics[width=1\linewidth]{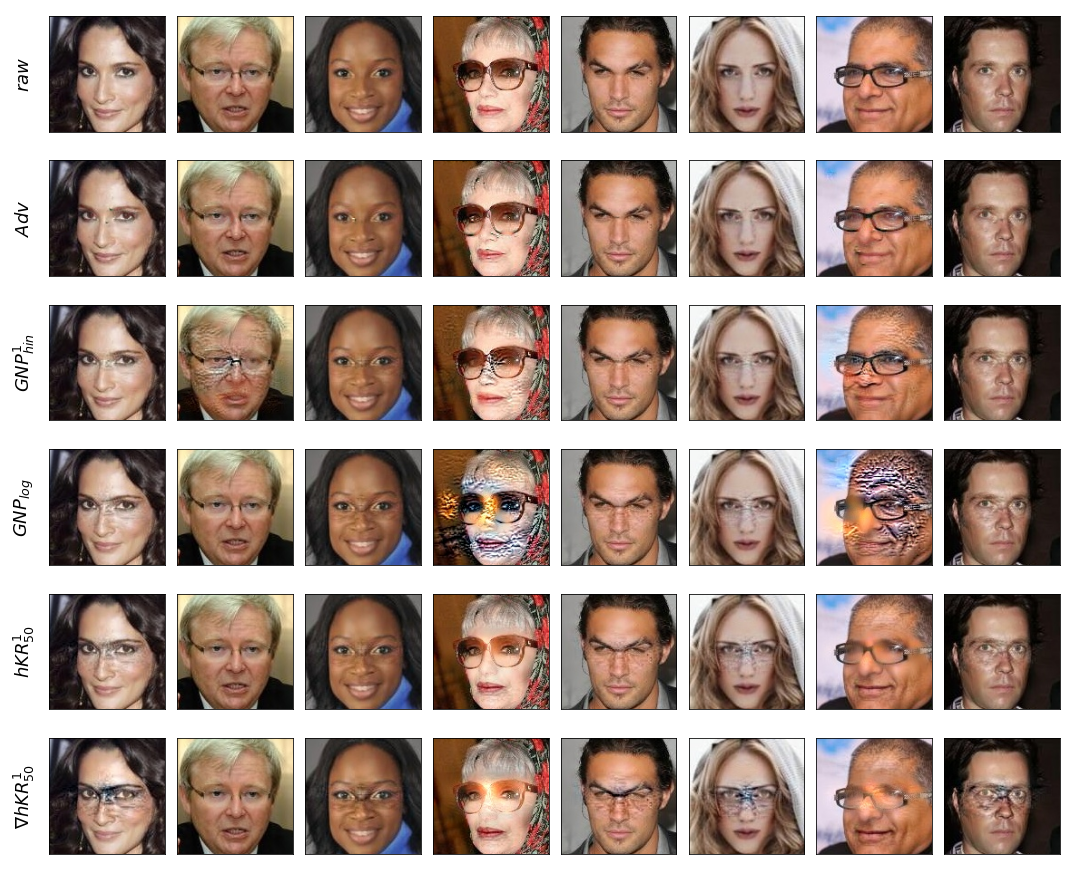}

\caption{Adversarial examples on CelebA dataset. The first row is the input image.}
\label{fig:Deep_fool_CelebA}
\end{figure*}

Moreover, Table \ref{tab:celeba_comparison} shows that, for the proposed solution, accuracies w.r.t. $\epsilon$ are similar regardless the attacks. This suggests that optimal attacks are the same, and they are in the direction of the gradient of the classifier. This confirms the intuition pointed out in section~\ref{sec:khr_properties} that optimal attack consists on traveling along the optimal transport map. In Figure~\ref{fig:Deep_fool_CelebA}, we compare adversarial images obtained with $l_2$ deepfool for the different models. The first row corresponds to the initial image. The following rows except the last one, are pictures obtained after attacks for the different models. We design the attacks to push the image just beyond the classification boundary (50/50).  For the Madry et al. approach ($Adv$) the noise is barely detectable. The noise is more visible with the gradient preserving approaches $GNP_{log}$ and $GNP^m_{hin}$ but it is still hard to interpret and sometimes meaningless. In contrast to the other approaches, the noise required to change the output class for the $hKR$ classifier is highly structured, and interpretable. For people without glasses, the attack adds noise around the eyes and at the top of the nose. For people with eyeglasses, the attack tends to erase the glasses around the eyes and at the top of the nose. 

The last row corresponds to the scheme proposed in Section~\ref{sec:khr_properties} $att(\hat{f},\textbf{x})\approx x-2*\hat{f}(\textbf{x})*\nabla_x \hat{f}(\textbf{x})$ where $\hat{f}$ is the $hKR^1_{50}$ model. Indeed, if $\nabla_x \hat{f}(\textbf{x})=1$ and $\hat{f}(\textbf{x})$ is the distance between the $\textbf{x}$ and its image with respect to the transport, we expect to have $att(\hat{f},\textbf{x})=tr(\hat{f},\textbf{x})$. The  shifts are similar to the attacks and even more meaningful. This confirms that attacks against $hKR$ can be interpreted as a transportation from a class to the other one and then requires to explicitly change the transport image in the opposite class. This suggests that attacks can explain classification for the $hKR$ model. Moreover, the gradient of this model allows to build this explanation without relying on time-consuming algorithms.

\begin{table}[]
    \centering

\begin{tabular}{|c|c|c|c|c|c|}
\hline
                          & $\epsilon$   & $Adv$ & $GNP_{hin}$ & $GNP_{log}$ & $hKR_{50}$\\ \hline
Base                      &   0   &   \textbf{98.04 }  &  97.2        &       94.51    &     96.07         \\ \hline
\multirow{2}{*}{$FGSM$} & $2$     &    \textbf{90.52 } &   86.79      &         81.40  &    87.44             \\ \cline{2-6} 
                          &  $5$  &  \textbf{ 82.96  } &     37.70    &      43.90     &     64.01                \\ \hline
\multirow{2}{*}{$l_2 GPD$}&  $2$            & 81.98    &     88.65    &    84.17       &\textbf{88.75  }           \\ \cline{2-6} 
                          &    $5$      &   61.35      &    34.90     &    48.02       &              \textbf{69.79  }     \\ \hline
\multirow{2}{*}{$l_2 CW$} &  $2$            & 32.14    &   80.80      &    79.02       &        \textbf{ 88.62 }         \\ \cline{2-6} 
                          &      $5$          &  13.17  &    29.46   &   32.81         &      \textbf{56.44     }    \\ \hline
\end{tabular}
    \caption{Robustness w.r.t. various attacks on CelebA dataset}
    \label{tab:celeba_comparison}
\end{table}

\section{Conclusion and future works}
\label{sec:conclusion}
This paper presents a novel classification framework and the associated deep learning process. Besides the interpretation of the classification task as a regularized optimal transport problem, we demonstrate that this new formalization has some valuable properties about error bounds and structural robustness regarding adversarial attacks. We also propose a systematic approach to ensure the 1-Lipschitz constraint of a neural network. This includes a state-of-the-art regularization algorithm and more precise constant evaluation for convolutional and pooling layers. Even if this regularization process can increase the computation time during learning, it doesn't impact inference. We developed an open source python library based on tensorflow for 1-Lipshitz layers and gradient preserving activation and pooling functions. This makes the approach very easy to implement and to use. \\

The experiment emphasizes the theoretical results and confirms that the classifier has good and predictable robustness to adversarial attacks with an acceptable cost on accuracy. We also show that our classifier forces adversarial attacks to explicitly modify the input. This suggest that our models can use adversarial attacks for explaining a prediction as it is done with counterfactual explanation in logic~\cite{DBLP:journals/corr/abs-1711-00399}.

In conclusion, we believe that this classification framework based on optimal transport is of great interest for critical problems since it provides both empirical and theoretical guarantees. 

\bibliographystyle{plain}
\bibliography{biblio} 
\newpage
\def\pd{\partial}
\def\sen{\mathop{\rm sen}\nolimits} 
\def\senh{\mathop{\rm senh}\nolimits}
\def\N{\mathbb{N}}
\def\Z{\mathbb{Z}}
\def\Q{\mathbb{Q}}
\def\R{\mathbb{R}}
\def\C{\mathbb{C}} 
\newtheorem {Lemma}[Proposition] {Lemma}
\newtheorem {Corollary}[Proposition]{Corollary}
\newtheorem {Remark}[Proposition]{Remark}
\newtheorem {Example}{Example}[section]
\newtheorem {Definition}{Definition}[section]
\newtheorem {Figure}{Figure}[section]
\newpage
\appendix

\section{Optimal transportation : discrete case}
\subsection{Optimal transport}
\label{secOptTransp}
When considering a limited number of samples for the two distribution, the computation of the Wasserstein distance 
can be solved through linear programming algorithms. In the balanced case, we have $X=\{x_1, \ldots, x_{2n}\}$ where $\{x_1,\ldots,x_n\}$ are sampled from $P_+$ and  $\{x_{n+1},\ldots,x_{2n}\}$ are sampled from $P_-$. We note $U=\{u_1, \ldots, u_{2n}\}$ the labels with $u_1,\ldots,u_n=1$ and $u_{n+1},\ldots,u_{2n}=-1$ and $C$ the $n\times n$ matrix cost function with $C_{i,j}=||x_i-x_{n+j}||$. The primal problem of the optimal transport is to find a transportation plan $\Pi$ (a $n\times n$ matrix) such as:
\begin{alignat}{2}
&\!\min_{\Pi}        &\qquad& \sum_{i,j \in n\times n}\Pi_{i,j}*C_{i,j}\\
&\text{subject to} &      & \Pi_{i,j} \geq 0,\\
&                  &      & \sum_i \Pi_{i,j} = \frac{1}{n},\sum_j \Pi_{i,j} = \frac{1}{n}.
\end{alignat}
The constraints enforce $\Pi$ to be a discrete joint probability distribution with the appropriate marginals as in the continuous case. The dual formulation for the discrete optimal transport problem is:
\begin{alignat}{2}
&\!\max_{F}        &\qquad& F.U^T\\
&\text{subject to} &      & \forall i,j \in n\times n, F_i-F_{n+j}\leq C_{i,j}
\end{alignat}
where $F$ is a $2n$ vector that is a discrete version of the function $f$ of Equation \ref{kantorovich}. The constraint on $F$ is the discrete counterpart of the 1-Lipschitz constraint. 

\subsection{Hinge regularized Optimal transport}
Similarly to the classical case, the discrete counterpart of the regularized Wasserstein distance is also a transportation problem which has the following formulation:
\begin{alignat*}{2}
&\!\min_{\Pi}        &\qquad& \sum_{i,j \in n\times n}\left[\Pi_{i,j}*C_{i,j}\right]-2\left(1-\sum_{i,j \in  n\times n}\left[\Pi_{i,j}\right]\right)\\
&\text{subject to} &      & \Pi_{i,j} \geq 0,\\
&                  &      & \frac{1}{n} \leq \sum_i \Pi_{i,j} \leq  \frac{1+\lambda}{n},\\
&                  &      & \frac{1}{n} \leq \sum_j \Pi_{i,j} \leq  \frac{1+\lambda}{n}.
\end{alignat*}
Roughly speaking, it allows to give more weight to the transportation of the closest pairs by admitting to deviate from the marginals with a tolerance that depends on $\lambda$. Since the closest pairs in the two distributions are the most difficult to classify, it illustrates why this formulation is more adequate for classification problems. 
The dual formulation of this transportation problem is a discrete counterpart of Equation \ref{eq:reg_OT} :
\begin{alignat*}{2}
&\!\max_{F}        &\qquad& \sum_{k =0}^{2n} \left[F_i*u_i -\lambda (0,1-F_i*u_i)_+\right]\\
&\text{subject to} &      & \forall i,j \in n\times n, F_i-F_{n+j}\leq C_{i,j}.
\end{alignat*}
We observe that the constraint in the dual problem are not affected by the new formulation and still corresponds to a the 1-Lipschitz constraint in the continuous case.

\section{Theorem proofs}
\label{appendix:proofs}
\subsection{Proof Theorem 1}
We denote as 
\begin{align}\label{eq:defwoth}
	f^*:=f_{\lambda}^*\in{\arg\min}_{f\in \text{Lip}_1(\Omega)}\mathcal{L}^{hKR}_\lambda(f)\ \ \text{and} \ \ \hat{f}_n:=\hat{f}_{n,\lambda}\in{\arg\min}_{f\in \text{Lip}_1(\Omega)}\hat{\mathcal{L}}^{hKR}_{\lambda,n}(f).
\end{align}
If we assume that \eqref{eq:M} is not true, then there exists $\textbf{x}\in \Omega$ such that $f^*(\textbf{x})>1+\text{diam}(\Omega)+\frac{\mathcal{R}(\psi)}{\inf(p,1-p)}$ or $f^*(\textbf{x})<-1-\text{diam}(\Omega)-\frac{L_1(\psi)}{\inf(p,1-p)}$. We suppose without loss of generality that the first inequality holds. If ${{\textbf{z}}}\in \Omega $ then the 1-Lipschitz condition in $f^*$ implies that $f^*({{\textbf{z}}})>1+\frac{L_1(\psi)}{1-p}$. Hence $(1-f^*)_+=0$ and
\begin{align*}
	L(f^*)&\leq \sup_{g\in Lip_1(\Omega)} L_2(g)- \lambda L_1(f^*)\\
	&=\sup_{g\in Lip_1(\Omega)}E_{X|Y=1}(g(X))-E_{X|Y=-1}(g(X))-E\left\{\lambda (1-Yf^*(X))_+\right\}\\
	&= L_2(\psi)-\lambda \{ pE_{X|Y=1}(1-f^*(X))_++(1-p)E_{X|Y=-1}(1+f^*(X))_+\}\\
	&\leq L_2(\psi)-\lambda \{ (1-p)E_{X|Y=-1}(1+f^*(X))\}\\
	&\leq  L_2(\psi)-\lambda \{ (1-p)E_{X|Y=-1}(2+\frac{L_1(\psi)}{1-p})\} \\
	&= L_2(\psi)-2\lambda(1-p) -\lambda \{ E_{X|Y=-1}({L_1(\psi)})\}=L_2(\psi)-2\lambda(1-p)-\lambda L_1(\psi) 
\end{align*}
Then $f^*$ can not be an optimal solution of the problem \eqref{eq:defwoth}. Then there exists some constant $M$ large enough, such that $f^*$ belongs to $\text{Lip}^M_1(\Omega):=\{f\in \text{Lip}_1(\Omega): || f||_{\infty}\leq M\}$ and not to $\text{Lip}_1(\Omega)$. Since the functional $\mathcal{L}^{hKR}_\lambda$ is convex and $\text{Lip}^M_1(\Omega)$ is compact in $\mathcal{C}(\Omega)$, we are able to make use of Ascoli-Arzela Theorem and conclude that there exists at least one function minimizing the expected loss. Furthermore the set of those functions is compact and convex.

\subsection{Proof Theorem 2}
\begin{Definition}
Let $\mu, \nu$ two positive measures in $\R^d$. The Kullback-Leibler divergence from $\mu$ to $\nu$ is defined as 
\begin{align}
KL(\mu |\nu )=\begin{cases} 
      \int \log(\frac{d\mu}{d\nu})d\mu-\int d\mu+\int d\nu & \text{if} \ \ \mu  << \nu\\
      \infty & otherwise 
   \end{cases}
\end{align}
\end{Definition}
\begin{Theorem}\label{Theo:generaldual}
Let $\phi_1, \phi_2:\Omega\rightarrow \bar{\R}$ be lower semicontinuous convex functions and $\mu, \nu\in \mathcal{P}(\Omega)$ be probability measures. Then for all $\epsilon>0$ the following equality holds
\begin{align}\label{eq:dual_app}
\begin{split}
	&\inf_{\pi \in\Pi_+(\mu, \nu)} \int{\phi_1(-\frac{d\pi_\textbf{x}}{d\mu}(\textbf{x}))d\mu}(\textbf{x})+\int{\phi_2(-\frac{d\pi_{{\textbf{z}}}}{d\nu}({{\textbf{z}}}))d\nu}({{\textbf{z}}})+\epsilon KL(\pi|e^{-\frac{c(\textbf{x},{{\textbf{z}}})}{\epsilon}}(d\mu\times d\nu))\\&= \sup_{f,g\in L^1(\Omega)}-\int_{\Omega}\phi_1^*(f(\textbf{x}))d\mu(\textbf{x})-\int_{\Omega}\phi_1^*(g({{\textbf{z}}}))d\nu({{\textbf{z}}})-\epsilon \int \left(e^{\frac{f(\textbf{x})+g({{\textbf{z}}})-c(\textbf{x},{{\textbf{z}}})}{\epsilon}}-e^{-\frac{c(\textbf{x},{{\textbf{z}}})}{\epsilon}}\right)d\mu d\nu.
	\end{split}
\end{align}
Furthermore if $\epsilon=0$ then
\begin{align}\label{eq:dual_app_zero}
\begin{split}
	&\inf_{\pi \in\Pi_+(\mu, \nu)}\int_{\Omega\times \Omega}c(\textbf{x},{{\textbf{z}}})d\pi(\textbf{x},{{\textbf{z}}}) + \int{\phi_1(-\frac{d\pi_\textbf{x}}{d\mu}(\textbf{x}))d\mu}(\textbf{x})+\int{\phi_2(-\frac{d\pi_{{\textbf{z}}}}{d\nu}({{\textbf{z}}}))d\nu}({{\textbf{z}}})\\&= \sup_{(f,g)\in \Phi(\mu, \nu)}-\int_{\Omega}\phi_1^*(f(\textbf{x}))d\mu(\textbf{x})-\int_{\Omega}\phi_1^*(g({{\textbf{z}}}))d\nu({{\textbf{z}}}).
	\end{split}
\end{align}
Where $\Pi_+(\mu, \nu)$ is the set of positive measures $\pi\in \mathcal{M}_+(\Omega\times \Omega)$ which are absolutely continuous  with respect to the joint measure $d\mu\times d\nu$, and $\Phi(\mu, \nu)$ consists of the pairs of functions $(f,g)\in L_1(\Omega)\times L_1(\Omega)$ such that $c(\textbf{x},{{\textbf{z}}})-f(\textbf{x})-g({{\textbf{z}}}) \geq 0$ $ d\mu\times d\nu -a.s.$.
\end{Theorem}

First we recall the Fenchel–Rockafellar Duality result, we use a weaker version given in Theorem 1.12 in \cite{Brez}
\begin{Proposition} \label{Fenchel–Rockafellar}
Let $E$ be a Banach space and $\Upsilon,\Psi:E\rightarrow \R \cup \{\infty \}$ be two convex functions, assume that there exist ${{\textbf{z}}}_0\in\text{dom}(\Psi)\cap \text{dom}(\Upsilon) $ such that $\Psi$ is continous in ${{\textbf{z}}}_0$. Then strong duality holds
\begin{align} \label{FC}
\inf_{a\in E}\left\lbrace \Upsilon(a)+\Psi(a)\right\rbrace=\sup_{b\in E^*}\left\lbrace -\Upsilon^*(-b)-\Psi^*(b)\right\rbrace
\end{align}
\end{Proposition} 
We identify the different elements of our problem with such of previous Proposition. 
\begin{itemize}
\item $E$ is the space of continuous functions in $\Omega\times \Omega$. Note that the set is bounded, hence $E^*$, by Riesz theorem, is the set of regular measures in $\Omega\times \Omega$.
\item If $\epsilon=0:$
\begin{align}
\Psi_0(u)&=\begin{cases} 
      0 & \text{if} \ \ u(\textbf{x},{{\textbf{z}}})\geq -c(\textbf{x},{\textbf{z}})\\
      \infty & \text{otherwise }
   \end{cases}\\ 
\Upsilon_0(u)&= \begin{cases} 
      \int \phi_1^*(-f(\textbf{x}))d\mu(\textbf{x})+\int \phi_2^*(-g({{\textbf{z}}})) d\nu({{\textbf{z}}})  & \text{if} \ \ u(\textbf{x},{\textbf{z}})=f(\textbf{x})+g({\textbf{z}})\\ 
      \infty & \text{otherwise } 
   \end{cases}
\end{align}
If $\epsilon>0$:
\begin{align}
\Psi_{\epsilon}(u)&=\epsilon \int \left(e^{\frac{u(\textbf{x},{{\textbf{z}}})-c(\textbf{x},{\textbf{z}})}{\epsilon}}-e^{-\frac{c(\textbf{x},{\textbf{z}})}{\epsilon}}\right)d\mu(\textbf{x}) d\nu({{\textbf{z}}})\\ 
 \Upsilon_{\epsilon}(u)&=\Upsilon_0(u)
\end{align}
\end{itemize}
Note that $ \Upsilon_{\epsilon}(u)=\Upsilon_0(u)$ could be non well defined, to avoid this situation we fix $\textbf{x}_0\in \Omega$ and consider $u(\textbf{x},{{\textbf{z}}})=(u(\textbf{x},{{\textbf{z}}}_0)-u({{\textbf{z}}}_0,{{\textbf{z}}}_0)/2)+u({{\textbf{z}}}_0,{{\textbf{z}}})-u({{\textbf{z}}}_0,{{\textbf{z}}}_0)/2)$. 
Now we compute the dual operators 
\begin{align*}
\Psi^*_{\epsilon}(-\pi)&=\sup_{u\in E}\left\lbrace  -\epsilon\int \left(e^{\frac{u(\textbf{x},{{\textbf{z}}})-c(\textbf{x},{\textbf{z}})}{\epsilon}}-e^{-\frac{c(\textbf{x},{\textbf{z}})}{\epsilon}}\right)d\mu(\textbf{x}) d\nu({{\textbf{z}}})-\int u(\textbf{x},{\textbf{z}})d\pi(\textbf{x},{{\textbf{z}}}) \right\rbrace\\
&=\sup_{u\in E}\left\lbrace -\epsilon\int \left(e^{\frac{u(\textbf{x},{\textbf{z}})-c(\textbf{x},{{\textbf{z}}})}{\epsilon}}-e^{-\frac{c(\textbf{x},{{\textbf{z}}})}{\epsilon}}\right)d\mu(\textbf{x}) d\nu({{\textbf{z}}})+\int u(\textbf{x},{{\textbf{z}}})d\pi(\textbf{x},{{\textbf{z}}}) \right\rbrace
\end{align*}
Now if $\pi$ were not absolutely continuous respect the joint measure $e^{-\frac{c(\textbf{x},{{\textbf{z}}})}{\epsilon}}d\mu\times d\nu$ then we would have a continuous function $u(\textbf{x},{{\textbf{z}}})=0$ $d\mu\times d\nu$ almost surely and such that $\int u(\textbf{x},{{\textbf{z}}})d\pi(\textbf{x},{{\textbf{z}}})\neq 0$. If we take the function $\lambda u({{\textbf{z}}},{{\textbf{z}}})$ and $\lambda$ tends to $\pm \infty$ we deduce that the supremum is $\infty $. Then suppose that $d\pi =m(\textbf{x},{{\textbf{z}}})e^{-\frac{c(\textbf{x},{{\textbf{z}}})}{\epsilon}}(d\mu\times d\nu)$.
\begin{align*}
\Psi^*_{\epsilon}(-\pi)&= \begin{cases} 
      \sup_{u\in E}\left\lbrace \epsilon\int \left(-e^{\frac{u(\textbf{x},{{\textbf{z}}})}{\epsilon}}+1 +\frac{u(\textbf{x},{{\textbf{z}}})}{\epsilon}m(\textbf{x},{{\textbf{z}}})\right) e^{-\frac{c(\textbf{x},{{\textbf{z}}})}{\epsilon}}d\mu(\textbf{x}) d\nu({{\textbf{z}}}) \right\rbrace & \text{if $d\pi =m e^{-\frac{c(\textbf{x},{{\textbf{z}}})}{\epsilon}}(d\mu\times d\nu)$.} \\
      \infty & \text{otherwise. } 
   \end{cases}\\
   &=  \epsilon KL(\pi|e^{-\frac{c(\textbf{x},{{\textbf{z}}})}{\epsilon}}(d\mu\times d\nu) ) & 
\end{align*}
With some similar calculation, we compute for $\epsilon=0$:
\begin{align*}
\Psi^*_{0}(-\pi)&= \begin{cases} 
     \int c(\textbf{x},{{\textbf{z}}})d\pi(\textbf{x},{{\textbf{z}}}) & \text{if $\pi$ is a positive measure.} \\
      \infty & \text{otherwise. } 
   \end{cases}
\end{align*}
Finally for $\Upsilon^*_{\epsilon}=\Upsilon^*_{0}$
\begin{align*}
\Upsilon^*_{\epsilon}(\pi)&=\sup_{u\in E,\  u(\textbf{x},{{\textbf{z}}})=f(\textbf{x})+g({{\textbf{z}}}) }\left\lbrace \int f(\textbf{x})+g({{\textbf{z}}}) d\pi(\textbf{x},{{\textbf{z}}})-\int \phi_1^*(-f(\textbf{x}))d\mu(\textbf{x}) -\int \phi_2^*(-g({{\textbf{z}}})) d\nu({{\textbf{z}}})\right\rbrace\\
 &=\sup_{u\in E,\  u(\textbf{x},{{\textbf{z}}})=f(\textbf{x})+g({{\textbf{z}}}) }\left\lbrace \int f(\textbf{x})d\pi_\textbf{x}(\textbf{x})-\int \phi_1^*(-f(\textbf{x}))d\mu(\textbf{x}) +\int g({{\textbf{z}}}) d\pi_{{\textbf{z}}}({{\textbf{z}}})-\int \phi_2^*(-g({{\textbf{z}}})) d\nu({{\textbf{z}}})\right\rbrace\\
  &=\sup_{f \in C(\Omega)}\left\lbrace \int f(\textbf{x})d\pi_\textbf{x}(\textbf{x})-\int \phi_1^*(-f(\textbf{x}))d\mu(\textbf{x})\right\rbrace +\sup_{g\in C(\Omega) }\left\lbrace \int g({{\textbf{z}}}) d\pi_{{\textbf{z}}}({{\textbf{z}}})-\int \phi_2^*(-g({{\textbf{z}}})) d\nu({{\textbf{z}}})\right\rbrace\\
  &=(I_1)+(I_2).
\end{align*}
We first consider  $(I_1)$. The same reasoning will hold for $(I_2)$. If $\pi_\textbf{x}$ were not absolutely continuous respect $\mu$ then reasoning as before we obtain $\infty$. Then $d \pi_\textbf{x}=\frac{d\pi_\textbf{x}}{d\mu}d\mu$ and
\begin{align*}
(I_1)&=\sup_{f \in C(\Omega)}\left\lbrace \int \left( -f(\textbf{x})\frac{d\pi_\textbf{x}}{d\mu}-\phi_1^*(f(\textbf{x})))\right)d\mu(\textbf{x})\right\rbrace\\
&=\int \left( \sup_{m}\left\lbrace -\frac{d\pi_\textbf{x}}{d\mu}m- \phi_1^*(m))\right\rbrace\right)d\mu(\textbf{x})=\int \phi_1(-\frac{d\pi_\textbf{x}}{d\mu})d\mu(\textbf{x})\\
(I_2)&=\int \phi_2(-\frac{d\pi_\textbf{x}}{d\nu})d\mu({{\textbf{z}}})
\end{align*}
Note that the inversion of the  supremum and  the integral  is guaranteed here since $(\textbf{x},m)\mapsto -m\frac{d\pi_\textbf{x}}{d\mu}(\textbf{x})+\phi_1^*(m)$ is lower semi-continuous and convex in $m$ and measurable in $(\textbf{x},m)$. Then it is a normal integrand, and we can apply Theorem 14.60 in \cite{RoWe}.\\
Then computing both in Equation \eqref{FC} we end with the following result
\begin{align*}
\inf_{ u(\textbf{x},{\textbf{{{\textbf{z}}}}})=f(\textbf{x})+g({\textbf{{{\textbf{z}}}}})\geq -c(\textbf{x},{{\textbf{z}}})}&\left\lbrace \int \phi_1^*(-f(\textbf{x}))d\mu(\textbf{x})+\int \phi_2^*(-g({{\textbf{z}}})) d\nu({{\textbf{z}}})\right\rbrace\\& = \inf_{ f(\textbf{x})+g({{\textbf{z}}})\leq c(\textbf{x},{{\textbf{z}}})}\left\lbrace \int\phi_1^*(f(\textbf{x}))d\mu(\textbf{x})+\int \phi_2^*(g({{\textbf{z}}})) d\nu({{\textbf{z}}})\right\rbrace\\
\\& = -\sup_{ f(\textbf{x})+g({{\textbf{z}}})\leq c(\textbf{x},{{\textbf{z}}})}\left\lbrace -\int \phi_1^*(f(\textbf{x}))d\mu(\textbf{x})-\int \phi_2^*(g({{\textbf{z}}})) d\nu({{\textbf{z}}})\right\rbrace
\end{align*}
\begin{align*}
\sup_{ \pi\in \mathcal{M}_+(\Omega\times \Omega)}&\left\lbrace  -\int c(\textbf{x},{{\textbf{z}}})d\pi(\textbf{x},{\textbf{{{\textbf{z}}}}})-\int \phi_2(-\frac{d\pi_\textbf{x}}{d\nu})d\mu({{\textbf{z}}})- \int \phi_1(-\frac{d\pi_\textbf{x}}{d\mu})d\mu(\textbf{x})\right\rbrace\\
=&-\inf_{ \pi\in \mathcal{M}_+(\Omega\times \Omega)}\left\lbrace\epsilon \int c(\textbf{x},{\textbf{{{\textbf{z}}}}})d\pi(\textbf{x},{\textbf{{{\textbf{z}}}}})+\int \phi_2(-\frac{d\pi_\textbf{x}}{d\nu})d\mu({{\textbf{z}}})+ \int \phi_1(-\frac{d\pi_\textbf{x}}{d\mu})d\mu(\textbf{x})\right\rbrace.
\end{align*}
\begin{align*}
\inf_{ u(\textbf{x},{\textbf{{{\textbf{z}}}}})=f(\textbf{x})+g({\textbf{{{\textbf{z}}}}})}&\left\lbrace \epsilon \int \left(e^{\frac{-f(\textbf{x})-g({\textbf{{{\textbf{z}}}}})-c(\textbf{x},{\textbf{{{\textbf{z}}}}})}{\epsilon}}-e^{-\frac{c(\textbf{x},{\textbf{{{\textbf{z}}}}})}{\epsilon}}\right)d\mu(\textbf{x}) d\nu({{\textbf{z}}})+\int\phi_1^*(-f(\textbf{x}))d\mu(\textbf{x})+\int \phi_2^*(-g({{\textbf{z}}})) d\nu({{\textbf{z}}})\right\rbrace\\ & = \inf_{f,g}\left\lbrace \epsilon \int \left(e^{\frac{f(\textbf{x})+g({\textbf{{{\textbf{z}}}}})-c(\textbf{x},{\textbf{{{\textbf{z}}}}})}{\epsilon}}-e^{-\frac{c(\textbf{x},{\textbf{{{\textbf{z}}}}})}{\epsilon}}\right)d\mu(\textbf{x}) d\nu({{\textbf{z}}})+\int\phi_1^*(f(\textbf{x}))d\mu(\textbf{x})+\int \phi_2^*(g({{\textbf{z}}})) d\nu({{\textbf{z}}})\right\rbrace\\
& = -\sup_{f,g}\left\lbrace -\epsilon \int \left(e^{\frac{f(\textbf{x})+g({\textbf{{{\textbf{z}}}}})-c(\textbf{x},{\textbf{{{\textbf{z}}}}})}{\epsilon}}-e^{-\frac{c(\textbf{x},{\textbf{{{\textbf{z}}}}})}{\epsilon}}\right)d\mu(\textbf{x}) d\nu({{\textbf{z}}})-\int\phi_1^*(f(\textbf{x}))d\mu(\textbf{x})-\int \phi_2^*(g({{\textbf{z}}})) d\nu({{\textbf{z}}})\right\rbrace\\
\end{align*}
\begin{align*}
\sup_{ \pi\in \mathcal{M}_+(\Omega\times \Omega)}&\left\lbrace-\epsilon KL(\pi|e^{-\frac{c(\textbf{x},{\textbf{{{\textbf{z}}}}})}{\epsilon}}(d\mu\times d\nu) )-\int \phi_2(-\frac{d\pi_\textbf{x}}{d\nu})d\mu({{\textbf{z}}})- \int \phi_1(-\frac{d\pi_\textbf{x}}{d\mu})d\mu(\textbf{x})\right\rbrace\\
=&-\inf_{ \pi\in \mathcal{M}_+(\Omega\times \Omega)}\left\lbrace\epsilon KL(\pi|e^{-\frac{c(\textbf{x},{\textbf{{{\textbf{z}}}}})}{\epsilon}}(d\mu\times d\nu) )+\int \phi_2(-\frac{d\pi_\textbf{x}}{d\nu})d\mu({{\textbf{z}}})+ \int \phi_1(-\frac{d\pi_\textbf{x}}{d\mu})d\mu(\textbf{x})\right\rbrace\\
\end{align*}
Proof of Theorem~\ref{Theo:dual}
With the same notation of Theorem \ref{Theo:generaldual}, it is enough to consider, $\mu=P_+$ $\nu=P_-$ and
\begin{align}
	\begin{split}\label{eq:coro:dual_general}
	\psi_1(s)&=\left\{ \begin{array}{lcc}
             p-s &   if  & s\in [p, p+\lambda p]\\
             \\ \infty &  else. &
             \end{array}
   \right.\\
	\end{split}\\
	\begin{split}\label{eq:coro:dual_general2}
	\psi_2(s)&=\left\{ \begin{array}{lcc}
             1-p-s &   if  & s\in [1-p, 1-p+\lambda (1-p)]\\
             \\ \infty &  else. &
             \end{array}
   \right.\\
	\end{split}
\end{align}
 Then for each $f\in L_1(d\mu), \ g\in L_1(d\nu)$
\begin{align*}
-\psi_1^*(f(\textbf{x}))&=-\sup_{s}	\{ -\psi_1(s)+f(\textbf{x})s\}=\inf_{s}	\{ \psi_1(-s)-f(\textbf{x})s\}=\inf_{s}	\{ \psi_1(s)+f(\textbf{x})s\}\\
&=\left\{ \begin{array}{lcc}
             f(\textbf{x}) &   if  & 1\leq f(\textbf{x})\\
             \\ f(\textbf{x})-p\lambda(1-f(\textbf{x})) &  else. &
             \end{array}
   \right.\\&=f(\textbf{x})-p\lambda(1-f(\textbf{x}))_+\\
-\psi_2^*(g(\textbf{{{\textbf{z}}}}))&= f(\textbf{{{\textbf{z}}}})-(1-p)\lambda(1-f(\textbf{{{\textbf{z}}}}))_+.
\end{align*}
Note that when $\lambda\geq 0$ the functions $r\mapsto h_1(r):=r-p\lambda(1-r)_+$ and $h_2(r):=r-(1-p)\lambda(1-r)_+$ are nondecreasing. Now if we denote as $J$ the right hand side of \eqref{eq:dual_app} then 
\begin{align*}
	J=\sup_{(f,g)\in \Phi(\mu, \nu)}\int_{\Omega}h_1(f({\textbf{x}}))d\mu({\textbf{x}})+\int_{\Omega}h_2(g({{\textbf{z}}}))d\mu({{\textbf{z}}}).
\end{align*}
We denote as $f^d$ the $d-$conjugate of $f$ defined as the function
 $$f^d(r):=\inf_{s\in\Omega}{\{ |r-s|-f(s)\}},$$   see for instance in~\cite{GaMc} for a suitable definition. It is clear that $f^{dd}\geq f$, and the equality holds if $f$ is a $d-$concave function since it is said that $f$ is $d-$concave if it is the $d$-conjugate of another function.  Hence using the nondecreasing condition of $h$ we get to
    \begin{align*}
	J=\sup_{f^{dd},f^d}\int_{\Omega}h_1(f^{dd}({\textbf{x}}))d\mu({\textbf{x}})+\int_{\Omega}h_2(f^{d}({{\textbf{z}}}))d\nu({{\textbf{z}}}).
\end{align*}
On the other side $f^d(r)=\inf_{s\in\Omega}{\{ |r-s|-f(s)\}}$ is a limit of a sequence of $1-$Lipschitz functions in $\Omega$, hence it belongs to $\text{Lip}_1(\Omega)$. Using the $1$-Lipschitz property and taking $r=s$ in the infimum leads to 
$$ -f^d(r)\leq  \inf_{s\in\Omega}{\{ |r-s|-f^d(s)\}}\leq -f^d(r).$$ 
This means that $f^{dd}=-f^d(r)$, hence we have that 
\begin{align*}
	J&=\sup_{(-f^d, f^d)}\int_{\Omega}h_1(f^{dd}({\textbf{x}}))d\mu({\textbf{x}})+\int_{\Omega}h_2(f^{d}({{\textbf{z}}}))d\mu({{\textbf{z}}}).\\
	&\leq \sup_{f\in \text{Lip}_1(\Omega)}\int_{\Omega}h_1(f^{dd}({\textbf{x}}))d\mu({\textbf{x}})+\int_{\Omega}h_2(-f({{\textbf{z}}}))d\nu({{\textbf{z}}})\leq J
\end{align*}
where the last inequality comes from the fact that if $f\in\text{Lip}_1(\Omega)$ then $(f,-f)\in \Phi(\mu, \nu)$.
\subsection{Proof Proposition 1}
\indent 
Even though the proof of Proposition~ \ref{coro_norm1} can be done following the frame of the proof of  Proposition 1 in \cite{gulrajani2017improved}, we have provided here   an easier proof  in order to make this document  self-content.  The proof of this Proposition requires some properties on the transport plan.
\begin{Definition}\label{def:c_ci}
A set $\Gamma\subset R^d\times \R^d$ is said to be $d$-cyclically monotone if for all $n\in \N$ and $\{(x_k,y_k)\}_{k=1}^{n} \subset \Gamma $ it is satisfied 
\begin{align}
	\sum_{k=1}^n c(x_k,y_k)\leq \sum_{k=1}^n c(x_{k+1},y_k), \ \ \text{assuming that  $n+1=1$}. 
\end{align}
It is said that a measure is  $d-$cyclically monotone if its support is  $d-$cyclically monotone.
\end{Definition}
In particular the optimal transference plan in Kantorovich problem for the cost $d$ is $d-$cyclically monotone, see Theorem 2.3 \cite{GaMc}. The same characterization holds for the optimal measures of \eqref{eq:dual_app}, this claim is proved in the following Lemma.
\begin{Lemma}\label{lem:d_mono}
The optimal measure $\pi$ of \eqref{eq:dual_app} is $d-$cyclically monotone for $d(\textbf{x},\textbf{z})=||\textbf{x}-\textbf{z} ||$.
\end{Lemma}
If $\pi$ were not $d-$cyclically monotone, in \cite{villani2008} it is built another measure $\tilde{\pi}$, with the same marginals as $\pi$, such that the value of $\int |\textbf{x}-\textbf{z}|d\pi(\textbf{x},\textbf{z})>\int |\textbf{x}-\textbf{z}|\tilde{\pi}(\textbf{x},\textbf{z})$. Computing this we deduce
\begin{align*}
\begin{split}
	\inf_{\pi \in\Pi^p_{\lambda}(\mu, \nu)}\int_{\Omega\times \Omega}|{\textbf{x}}-{{\textbf{z}}} |d\pi + \pi_{\textbf{x}}(\Omega)+\pi_{{\textbf{z}}}(\Omega)-1> \inf_{\tilde{\pi} \in\tilde{\pi}^p_{\lambda}(\mu, \nu)}\int_{\Omega\times \Omega}|{\textbf{x}}-{{\textbf{z}}} |d\tilde{\pi} + \tilde{\pi}_{\textbf{x}}(\Omega)+\tilde{\pi}_{{\textbf{z}}}(\Omega)-1.
	\end{split}
\end{align*}
Hence $\pi$ cannot be optimal. \\
We replicate this construction on order to build this proof as self content as possible.\\ 
{If $P_+$ and $P_-$ are discrete probabilities}. Then $P_+=\sum_{k=1}^n u_k\delta_{\textbf{x}_k}$ and $P_-=\sum_{j=1}^n v_j\delta_{\textbf{z}_j}$ then the optimal measure has the form: 
\begin{align}
 \frac{1}{n}\sum_{k,j=1}^n \pi_{k,j}\delta_{\textbf{x}_k,\textbf{z}_j}
\end{align}
If it is not $d$-cyclically monotone then there exist $N\in \N$ and $\{(\textbf{x}_{k_i},\textbf{z}_{k_i})\}_{i=1}^N \subset \text{supp}(\pi)$ such that:
\begin{align*}
\sum_{i=1}^{N}||\textbf{x}_{k_i}-\textbf{z}_{k_{i+1}} ||<\sum_{i=1}^{N}||\textbf{x}_{k_i}-\textbf{z}_{k_i} ||, \ \ \text{assuming that  } k_{N+1}=k_1.
\end{align*}
Let $a:=\inf_{i=1,\dots, N}\{ \pi_{k_i,k_i} \}>0$. And let's define $\tilde{\pi}$ as 
\begin{align*}
\tilde{\pi}:=\pi+\frac{1}{n}\sum_{i=1}^n \left( \delta_{\textbf{x}_{k_{i}},\textbf{z}_{k_{i+1}}}-\delta_{\textbf{x}_{k_{i}},\textbf{z}_{k_{i}}}\right).
\end{align*}
Then $$\tilde{\pi}(A\times \Omega)=\pi(A\times \Omega)+\frac{1}{n}\sum_{i=1}^n \left( \delta_{\textbf{x}_{k_{i}}}(A)-\delta_{\textbf{x}_{k_{i}}}(A)\right)=\pi(A\times \Omega).$$
And the same holds with $ ( \Omega\times B) $ and the other marginal, and also it satisfied that
\begin{align*}
\frac{1}{n}\sum_{k,j=1}^n | \textbf{x}_k-\textbf{z}_j|\tilde{\pi}_{k,j}< \frac{1}{n}\sum_{k,j=1}^n || \textbf{x}_k-\textbf{z}_j||\pi_{k,j}.
\end{align*}
Hence $\tilde{\pi}$ is the searched measure in the discrete case.
\\
{$\Pi^p_{\lambda}(\mathcal{S}, \mathcal{T})$ is sequentially compact respect the weak convergence denoted * of measures if both $\mathcal{S},\ \mathcal{T}$ are also.}
Because of the compactness of $\Omega \times \Omega$, we only have to check that the set is bounded in total variation. But this is straightforward because for each $\pi \in \Pi^p_{\lambda}(P_+, P_-) $ it is satisfied $|\pi|(\Omega\times \Omega)\leq (p+p\lambda)(p+p\lambda)$.
\\
{If $P_+$ and $P_-$ are general probabilities}. Let $X^+_1, \dots, X^+_n$ and $Z^+_1, \dots, Z^+_n$ be sequences of independent random variables with law $P_+$ and $P_-$. And let $P^+_n, \ P_-^n$ be the associated empirical measures. Buy using the strong law of large numbers we deduce that $P^+_n\rightarrow P_+$ and $ P_-^n\rightarrow P_-$ with probability one. 
\\ Now let $\pi_n$ be the corresponding optimal measure for $P^+_n, \ P_-^n$, then there exist a measure $\pi$ such that $\pi_n\rightharpoonup^* \pi$. It means that for each continuous and bounded function $f$ in $\Omega\times \Omega$ we get
\begin{align*}
\int f d\pi_n \longrightarrow \int f d\pi.
\end{align*}
Since the norm $(\textbf{x},\textbf{z})\mapsto ||\textbf{x}-\textbf{z} ||$ is continuous and bounded, once again because $\Omega$ is compact, we derive that 
\begin{align*}
\int||{\textbf{x}}-{{\textbf{z}}} ||d\pi_n + \pi_{\textbf{x}_n}(\Omega)+\pi_{{\textbf{z}}_n}(\Omega)-1 \longrightarrow \int||{\textbf{x}}-{{\textbf{z}}} ||d\pi + \pi_{\textbf{x}}(\Omega)+\pi_{{\textbf{z}}}(\Omega)-1
\end{align*}
Finally it is known that if a sequence of measures is $d$-cyclically monotone and converges weak* to another measure, then it is also $d$-cyclically monotone. This concludes the proof.

The proof of Proposition~\ref{coro_norm1} is achieved as follows.
The assumption of $d$-cyclically monotone involves that in particular $g(\textbf{x})-g(\textbf{z})=||\textbf{x}-\textbf{z} ||$ $\pi$-a.s. for some function $g$.  Then for the balanced case
\begin{align*}
\begin{split}
	&\int (g-1)d\pi_x- \int (g+1)d\pi_z+2\\&= \sup_{f\in \text{Lip}_1(\Omega)}\int_{\Omega}f(dP_+-dP_-)-\lambda\left( \int_{\Omega}(1-f)_+dP_+ +\int_{\Omega}(1+f)_+dP_-\right).
	\end{split}
\end{align*}
Then we split $(g-1)=(g-1)\mathds{1}_{g-1\geq 0}+(g-1)\mathds{1}_{g-1< 0}$ and 
\begin{align*}
\begin{split}
	&\int (g-1)d\pi_x+1\\&=(1+\lambda)\int (g-1)\mathds{1}_{g-1\geq 0} dP_++\int (g-1)\mathds{1}_{g-1< 0} dP_+=\int (g-1)- \lambda (1-g)_+ d P_+.
	\end{split}
\end{align*}
Doing the same with $P_-$, we deduce that this $g$ is optimal and  $g(\textbf{x})-g(\textbf{z})=||\textbf{x}-\textbf{z} ||$ $\pi$-a.s. for the optimal measure $\pi$.
As a consequence of such observations, following exactly the same arguments of the proof of Proposition 1 in \cite{gulrajani2017improved}, note that the key is  $g(\textbf{x})-g(\textbf{z})=||\textbf{x}-\textbf{z} ||$ $\pi$-a.s. which comes from what follows. \\
\indent
Let $f^*$ be the optimal of Lemma \ref{lem:d_mono}, $\textbf{x}$ be a differentiable point of $f^*$. By assumption, the density property implies that $\pi(\textbf{x}=\textbf{z})=0$, and then with probability one, there exist $\textbf{z}$ such that $f^*(\textbf{x})-f^*(\textbf{z})=||\textbf{x}-\textbf{z} || $ and both points are different $\textbf{x}\neq \textbf{z}$. For each $t\in[0,1]$ let $\textbf{x}_t=(1-t) \textbf{x}+t\textbf{z}$ and the path $\sigma:[0,1]\rightarrow \R$ defined as $\sigma(t):=f^*(\textbf{x}_t)-f^*(\textbf{x})$. The proof is split in two steps;
\\
\textbf{Step 1} ($\sigma(t)=||\textbf{x}_t-\textbf{z} ||=t|| \textbf{x}-\textbf{z}||$)\\
First of all we realize that for each $s,t\in [0,1]$
\begin{align*}
 |\sigma(t)-\sigma(s)|=| f^*(\textbf{x})-f^*(\textbf{x}_s) |\leq ||\textbf{x}_t-\textbf{x}_s||\leq | t-s|||\textbf{x}-\textbf{z} ||.
\end{align*}
Actually if we consider $t\in [0,1]$ then 
\begin{align*}
\sigma(1)-\sigma(0)&\leq \sigma(1)-\sigma(t)+\sigma(t)-\sigma(0)\\
&\leq (1-t) || \textbf{x}-\textbf{z}||+\sigma(t)-\sigma(0)
\\&\leq (1-t) || \textbf{x}-\textbf{z}||+t|| \textbf{x}-\textbf{z}||=|| \textbf{x}-\textbf{z}||=\sigma(1)-\sigma(0)
\end{align*}
And the inequalities become equalities and because $\sigma(0)=0$ we conclude $\sigma(t)=t|| \textbf{x}-\textbf{z}||$.
\\
\textbf{Step 2} (There exists some unitary vector $\textbf{v}$ such that $|(\partial f^* /\partial \textbf{v})(\textbf{x})|= 1$)\\
The candidate is $\textbf{v}=\frac{\textbf{z}-\textbf{x}}{||\textbf{x}-\textbf{z} ||}$, and lets compute the partial derivative
\begin{align*}
\frac{\partial f^* }{\partial \textbf{v}}(\textbf{x})&=\lim_{h\rightarrow 0}\frac{f(\textbf{x}+h\textbf{v})-f(\textbf{x})}{h}\\
&=\lim_{h\rightarrow 0}\frac{\sigma(\frac{h}{||\textbf{x}-\textbf{z} ||})}{h}=1.
\end{align*}
Then for each differentiable point $x$ of $f^*$ there exists an unitary vector $\textbf{v}$ such that $|\partial f^* /\partial \textbf{v}(\textbf{x})|= 1$. Then by creating an orthonormal base such that $\textbf{v}$ belongs to it we can deduce that $||\nabla f^*(\textbf{x})||=1$. And this event occurs with almost surely because of Rademacher Theorem.

\subsection{Proof Proposition 2}
As a direct consequence of Theorem \ref{Theo:dual} we derive the next equality 
 \begin{align}\label{eq:proofcoro}
\begin{split}
	&\inf_{\pi \in\Pi_{\lambda}(P_+, P_-)}\int_{\Omega\times \Omega}\left(\frac{1}{\epsilon}| {\textbf{x}}-{\textbf{{{\textbf{z}}}}} |-2\right)d\pi  +2\\&= \sup_{f\in \text{Lip}_{1/\epsilon}(\Omega)}\int_{\Omega}f(dP_+-dP_-)-\frac{\lambda}{2} \left( \int_{\Omega}(1-f)_+dP_+ +\int_{\Omega}(1+f)_+dP_-\right).
	\end{split}
\end{align}
We denote as $I$ the left hand side of \eqref{eq:proofcoro} and $\Pi(P_+, P_-)$ the set of measures with marginals $P_+, P_-$. Now using the hypothesis \eqref{coro:hipotesis} we derive the next inequality  
\begin{align*}
	I=\inf_{\pi \in\Pi(P_+, P_-)}\int_{\Omega\times \Omega}\left(\frac{1}{\epsilon}|{\textbf{x}}-{{\textbf{z}}} |-2\right)d\pi+2= \frac{1}{\epsilon}\mathcal{W}(P_+, P_-).
\end{align*}
Since { $\text{Lip}_{1/\epsilon}\mathcal{W}(P_+, P_-)=\sup_{f\in \text{Lip}_{1/\epsilon}(\Omega)}\int_{\Omega}f(dP_+-dP_-)$}, we denote as $\psi_{\epsilon}\in \text{Lip}_{1/\epsilon}(\Omega)$ the function where the supremum is achieved. Hence we derive the following inequality
\begin{align*}
  \frac{1}{\epsilon}\mathcal{W}(P_+, P_-)&= \int_{\Omega}f_{\lambda}(dP_+-dP_-)-\lambda \left( \int_{\Omega}(1-f_{\lambda})_+dP_+ +\int_{\Omega}(1+f_{\lambda})_+dP_-\right)\\
	&\leq \int_{\Omega}\psi_{\epsilon} (dP_+-dP_-)-\lambda \left( \int_{\Omega}(1-f_{\lambda})_+dP_+ +\int_{\Omega}(1+f_{\lambda})_+dP_-\right)\\
	&=\frac{1}{\epsilon}\mathcal{W}(P_+, P_-) -\lambda \left( \int_{\Omega}(1-f_{\lambda})_+dP_+ +\int_{\Omega}(1+f_{\lambda})_+dP_-\right).
\end{align*}
Then $ \int_{\Omega}(1-f_{\lambda})_+dP_+ +\int_{\Omega}(1+f_{\lambda})_+dP_-=0$ and the first assert of the proof is completed. The second assertion   is a straightforward consequence of the previous one.

\section{Lipshitz constant for convolutional networks}

\subsection{Enforcing 1-Lipschitz dense layer}
A neural network is a composition of linear and non-linear function. Let's study first a multilayer perceptron is defined as follows~:
$$f(x)=\phi_k(W_k.(\phi_{k-1}(W_{k-1}\ldots\phi_1(W_1.x))).$$
We name $L(f)$ the Lipschitz constant of a function $f$. As a composition of functions, the Lipschitz constant of a multilayer perceptron is upper bounded by the product of the individual Lipschitz constants: 
$$L(f)\leq L(\phi_k)*L(W_k)*L(\phi_{k-1})*L(W_{k-1})*\ldots*L(\phi_1)*L(W_1.x).$$
The most common activation functions such as ReLU or sigmoid are 1-Lipschitz. Thus, we can ensure that a perceptron is at most 1-Lipschitz by ensuring each dense layer $W_k$ is 1-Lipschitz. Given a linear function represented by an $n\times m$ matrix $W$, it is commonly admitted that:

\begin{equation}
\label{norm}
    L(W)=||W|| \leq ||W ||_F \leq \underset{ij}{max}(|W_{ij}|)*\sqrt{nm}
\end{equation}
where $|| W||$ is the spectral norm, and  $|| W||_F$ is the Frobenius norm. The initial version of WGAN \cite{Arjovsky2017} clips the weights of the networks. However, this is a very crude way to upper-bound the 1-Lipschitz (see equation \ref{norm}). Normalizing by the Frobenius norm have also been proposed in  \cite{SalimansK16}. In this paper, we use spectral normalization as proposed in \cite{Miyato2018SpectralNF}. At the inference step, we normalize the weights of each layer by dividing the weight by the spectral norm of the matrix:
$$W_s=\frac{W}{||W||}.$$
Even if this method is more computationally expensive than Frobenius normalization, it gives a finer upper bound of the 1-Lipschitz constraint of the layer.
The spectral norm is computed by iteratively evaluating the largest singular value with the power iteration algorithm \cite{GOLUB200035}. This is done during the forward step and taken into account for the gradient computation.

\subsection{Enforcing 1-Lipschitz convolutional layer}
\label{sec:ConvLayer1Lip}
\label{sec:ConvLayerMatrix}
In this section we will show that enforcing convolution kernels to 1-lispchitz is not enough for ensuring the 1-lipschitz property of convolutional layers, and will propose two normalization factors. 
Notations: We consider a Convolutional layer with an input feature map $X$ of size $(c,w,h)$, and $L$ output channels obtained with kernels $W=\{W_l\}_{l\in[0,L[}$ of odd size $(c,k,k)$, i.e. $k=2*\bar{k}+1$. Considering the classical \textit{same} configuration which output size is $(L,w,h)$, we use the following matrix notations of the convolution $Y=W \ast X$:
\begin{itemize}
    \item $\widetilde{X}$ the zero padded matrix of $X$ of size $(c,w+k-1,h+k-1)$
    \item $\bar{W}$ the vectorized matrix of weights of size $(L,c.k^2)$
    \item $\bar{X}$ a matrix of size $(c.k^2,w.h)$, a duplication of the input $\widetilde{X}$, where each column $j$ correspond to the $c.k^2$ inputs in $\widetilde{X}$ used for computing a given output $j$

    \item $\bar{Y} = \bar{W}.\bar{X}$ the vectorized output of size $(L,w.h)$
\end{itemize}

Given two outputs $X_1$ and $X_2$,  we can compute an upper bound of convolutional layer lipschitz constant (Eq.~\ref{eq:CnnMatrix}). 

\begin{equation} 
\begin{split}
||Y_1-Y_2||^2 &= ||\bar{Y_1} - \bar{Y_2}||^2 \leq ||\bar{W}||^2.||\bar{X_1}-\bar{X_2}||^2  \\
&\leq \Lambda^2.||W||^2.||X_1-X_2||^2
\end{split} 
\label{eq:CnnMatrix}\end{equation}
The coefficient $\Lambda^2$ can be estimated, as in~\cite{cisse_parseval_2017}, by the maximum number of duplication of the input matrix $\widetilde{X}$ in  $\bar{X}$: each input can be used at most in $k^2$ positions. But since within  $\bar{X}$, part of the values come from the zero padded zones in $\widetilde{X}$, and have no influence on $||Y_1-Y_2||^2$, we propose a tighter estimation of $\Lambda$, computing the average duplication factor of non zero padded value in  $\bar{X}$.

\begin{figure}[htp]
    \centering
    \includegraphics[width=7cm]{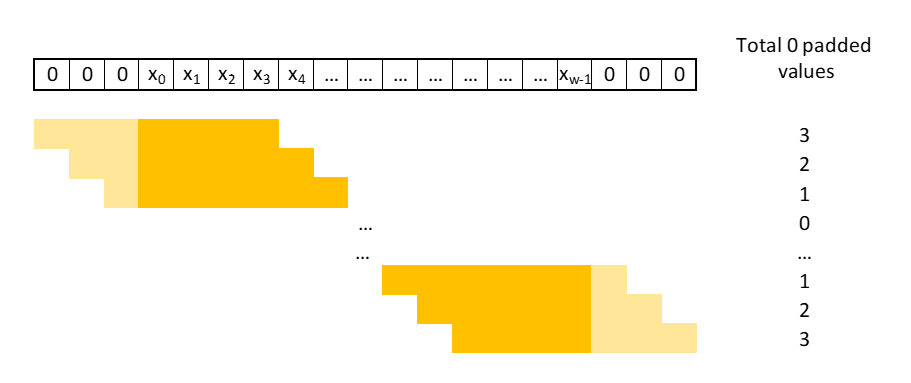}
    \caption{Zero padded elements in a 1D convolution with $k=7 (\bar{k} = 3)$ }
    \label{fig:conv1DNoStride}
\end{figure}

For a 1D convolution (see Fig.~\ref{fig:conv1DNoStride}), the number of zero values in the $\bar{k}$ first columns of $\bar{X}$ (symmetrically on the $\bar{k}$ last columns) is $(\bar{k},\bar{k}-1,...,1)$. So the number of zero padded values is $k.w-2*\sum_{t=1}^{\bar{k}}t=k.w-\bar{k}(\bar{k}+1)$.

We propose to use Eq.~\ref{eq:ConvApproxLip_app} as a tighter normalization factor\footnote{this factor Eq.~\ref{eq:ConvApproxLip_app} does not lead to a strict upper bound of the lipschitz constant, since particular matrix with high value on the center and low values on borders won't satisfy the inequality~(\ref{eq:CnnMatrix})}.
\begin{equation}
\Lambda=\sqrt{\frac{(k.w-\bar{k}.(\bar{k}+1)).(k.h-\bar{k}.(\bar{k}+1))}{h.w}}
\label{eq:ConvApproxLip_app}
\end{equation}

\subsection{Convolution layers with zero padding and stride }
\label{sec:convStride}
Convolution layers are sometimes used with stride (as in Resnet layers~\cite{he_deep_2015}) to reduce the computation cost of these layers\footnote{main drawback with stride is that each point in the input feature map has not the same number of occurrences}. Given a stride $(s,s)$, the output layer size of the layer will be $(wo,ho)$ such as $w=s.wo+rw$ and $h=s.ho+rh$. We also introduce $\alpha=\lceil \frac{k}{s} \rceil$ the maximum number of overlapping stride positions.
As in previous section, we can build a matrix $\bar{X}$ of size $(c.k^2,wo.ho)$, as a duplication of  $\widetilde{X}$. The maximum duplication factor of an element of $\widetilde{X}$ in  $\bar{X}$ is $\Lambda^2=\alpha^2$.

As in section \ref{sec:ConvLayerMatrix}, we can compute a tighter factor using the average duplication factor of input in $X$, by computing the number of non-zero-padded values used in $\bar{X}$. We introduce $\bar{\alpha},\bar{\beta}$ such as $\bar{k}=\bar{\alpha}.s+\bar{\beta}$.

\begin{figure}[htp]
    \centering
    \includegraphics[width=10cm]{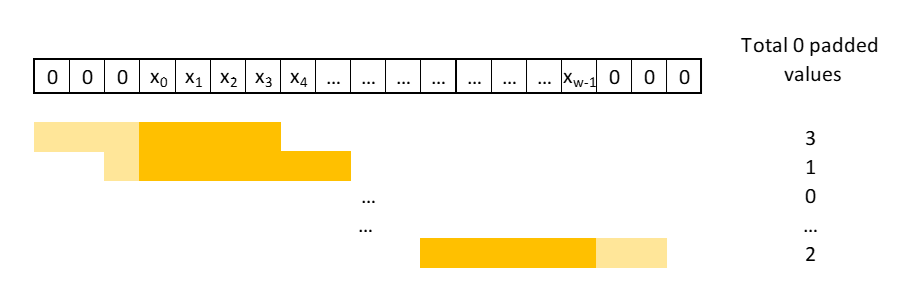}
    \caption{Zero padded elements in a 1D convolution with stride: $k=7$  $(\bar{k} = 3)$, and $s=2$}
    \label{fig:conv1DWithStride}
\end{figure}

For a 1D convolution  (see Fig.~\ref{fig:conv1DWithStride}), the number of zero values in the first columns of $\bar{X}$ is $(\bar{k},\bar{k}-s,...,\bar{\beta})$. So the number of zero padded values on the left side is $zl=\sum_{t=0}^{\bar{\alpha}}(\bar{k}-t.s)=(\bar{\alpha}+1)\bar{k}-s.\frac{\bar{\alpha}(\bar{\alpha}+1)}{2}=\frac{(\bar{\alpha}+1)(\bar{\alpha}s+2\bar{\beta})}{2}$.

On the right side (last columns), we introduce 
$\gamma_w = argmax\{\gamma=w-1-i.s,  \text{ such as } i>=0  \text{ and }\gamma\leq \bar{k}\}$ i.e. $\gamma_w = w-1-s.\lceil{\frac{w-1-\bar{k}}{s}} \rceil$.  $\gamma_w$ represents the first half-kernel to include the last element of the line. We also introduce $\alpha_w,\beta_w$ such as $\gamma_w = \alpha_w.s+\beta_w$. The number of zero values in the last columns is $(\bar{k}-\gamma_w,\bar{k}-\gamma_w+s,...,\bar{k}-\gamma_w+\alpha_w.s)$, i.e.  $zr_w=\sum_{t=0}^{\alpha_w}(\bar{k}-\gamma_w+t.s)=(\alpha_w+1)(\bar{k}-\gamma_w+\frac{s.\alpha_w}{2})$.

For the matrix $\bar{Y}$ the average duplication factor for a value of the input $X$ is $\frac{(k.wo-zl-zr_w).(k.ho-zl-zr_h)}{h.w}$

We propose to use Eq.~\ref{eq:ConvStrideApproxLip} as a tighter normalization factor\footnote{As in previous section,  this factor is not an upper bound of the lipschitz constant}\footnote{in case of stride $s=1$, we have $\bar{\alpha} = \bar{k}$, $\bar{\beta} = 0$, $\gamma_w= \alpha_w=\bar{k}$ and $\beta_w=0$. So we can retrieve $zl+zr_w=\frac{\bar{k}.(\bar{k}+1)}{2} + \frac{\bar{k}.(\bar{k}+1)}{2}=\bar{k}.(\bar{k}+1)$ }.
\begin{equation}
\Lambda=\sqrt{\frac{(k.wo-zl-zr_w).(k.ho-zl-zr_h)}{h.w}} \label{eq:ConvStrideApproxLip}
\end{equation}

\subsubsection{Pooling layers}
By definition, the max pooling layer is 1-lipschitz, since $||max(X_1)-max(X_2)||\leq ||X_1-X_2||$.

Considering average pooling layer with a averaging size of $po$, and a stride of $s$. Since a mean is equivalent to a convolution with the matrix $\frac{1}{po^2}\mathbbm{1}_{po\times po}$.
The average pooling layer is equivalent to a convolution with stride (sec \ref{sec:convStride}). Introducing $\alpha=\lceil \frac{po}{s} \rceil$, which is $1$ in the common case where $s=po$. So an upper bound of lipschitz constant for  the average pooling layer is  $\Lambda.||W||=\frac{\alpha}{po}$

\begin{table}[]
    \centering
    \begin{tabular}{|c|c|c|c|}
    \hline
       Layer type  & Parameters &\shortstack{Upper lip  \\constant}   & Thighter Lip estimation \\ \hline
       Dense & & $||W||$ & \\ \hline
       Convolution wo stride& \shortstack{kernel size $(k,k)$\\ $k=2\bar{k}+1$} & $k.||W||$ & $\sqrt{\frac{(k.w-\bar{k}.(\bar{k}+1)).(k.h-\bar{k}.(\bar{k}+1))}{h.w}}.||W||$\\\hline
       Convolution with stride &  \shortstack{kernel size $(k,k)$\\ stride $(s,s)$} & $\lceil \frac{k}{s} \rceil.||W||$ & $\sqrt{\frac{(k.wo-zl-zr_w).(k.ho-zl-zr_h)}{h.w}}.||W||$\\\hline
       MaxPoolig & & $1$ & \\\hline
       AveragePooling & \shortstack{averaging size $po$\\ stride $s$} &$\lceil \frac{po}{s} \rceil.\frac{1}{po}$ & \\ \hline
    \end{tabular}
    \caption{Main }
    \label{tab:wasserstein_estimation}
\end{table}

\subsection{Gradient norm preserving and general architecture}
\label{app:normpreserving}
As proven Sections \ref{sec:khr_properties} and , the optimal function $f^*$ with respect to Equation \ref{eq:reg_OT}, verifies $||\nabla f^* ||=1$ almost surely.
In \cite{gulrajani2017improved}, the authors propose to add a regularization terms with respect to the average gradient norm with respect to inputs in the loss function. However, the estimation of this value is difficult and a regularization term doesn't guarantee the property. In this paper, we apply the approach described in \cite{pmlr-v97-anil19a}, based on the use of specific activation functions and a normalization process of the weights. 
Three norm preserving activation functions are proposed:
\begin{itemize}
    \item \textbf{MaxMin} : order the vector by pairs.
    \item \textbf{GroupSort} : order the vector by group of a fixed size.
    \item \textbf{FullSort} : order the vector.
\end{itemize}
These function are vector-wise rather than element-wise. We also propose the activation  \textbf{ConstPReLU}, a PReLU \cite{He_2015} activation function complemented by a constraint such that $|\alpha|\leq 1$ ($\alpha$ the learnt slope). This last function is norm preserving only when $|\alpha|=1$ (linear, or absolute value function), but being computed element wise, it is then more efficient for convolutional layers outputs.

Given a vector $v$ of size $k$ the P-norm pooling is defined in \cite{boureau2010} as follows :
$$
Pool_{P-norm}(v)=\left(\frac{1}{k}\sum_{i=1}^k v_i^p P\right)^{\frac{1}{P}}
$$

Concerning gradient norm preserving linear layers, a weight matrix $W$ is norm preserving if and only if all the singular values of $W$ are equals to $1$. In \cite{pmlr-v97-anil19a}, the authors propose to use the Björk Orthonormalization algorithm \cite{bjorck71Ortho}. The Björk algorithm compute the closest orthonormal matrix by repeating the following operation :
\begin{equation}
    W_{k+1}=W_k(I+\sum_{i=1}^p(-1)^p \Spvek{-\frac{1}{2};p} Q_k^p)
\end{equation}
where $Q_k=1-W^T_kW_k$ and $W_0=W$.
This algorithm is fully differential, and as for spectral normalization, it is applied during the forward inference, and taken into account for back-propagation.
\subsection{Robustness bounds}
\label{sec:robustness_bounds}

Given a 1-lipschitz neural network $g$ and $N$  functions compose one 1-lipschitz dense layer with a single output $g_i$. We consider the multi-outputs neural network $f=[g_i\circ g]_{i\in[0,N[}$, and denote $f_i=g_i\circ g$.\\

For a given input $x$ of label $t$, we denote
\[
M_f(x) = max(0,f_t(x)-max_{i\neq t}(f_i(x))
\]

\begin{Theorem}[Adversarial Perturbation Robustness Condition under Lp Norm]
If $M_f(x)>2.\epsilon$ where $f=[g_i\circ g]_i$ is a concatenation of 1-lipschitz neural network under the $L_p$ norm. Then x is robust to any input perturbation $\Delta x$ with $||\Delta x||_p< \epsilon$
\end{Theorem}

Proof:
Suppose $x$ well classified of class $t$, such that $M_f(x) > 2\epsilon$.
We have 
\begin{equation*}
\forall i \neq t, f_t(x)-f_i(x) \geq M_f(x) > 2\epsilon
\end{equation*}

Given $\Delta x$ such that $||\Delta x||_p< \epsilon$, and $x'=x+\Delta x$.

Since $g_i$ and $g$ are 1-lipschitz, for all $i$, we have:
\begin{equation*}
|\Delta y_i|^p=|g_i\circ g(x')-g_i\circ g(x)|^p\leq ||g(x')-g(x)||_p^p \leq ||\Delta x||_p^p < \epsilon^p
\end{equation*}

So,
\[
|\Delta y_t|^p+|\Delta y_n|^p< 2.\epsilon^p
\]

\begin{equation*}
g(\frac{|\Delta y_n|+|\Delta y_t|}{2})\leq g(|\Delta y_t|) \iff  \frac{(|\Delta y_n| + \Delta y_t|)^p}{2^{p-1}} \leq|\Delta y_t|^p+|\Delta y_n|^p < 2.\epsilon^p
\end{equation*}

So, $\forall n\neq t$
\begin{equation*}
y'_t - y'_n = y_t -y_n + \Delta y_t - \Delta y_n \geq M_f(x) - (|\Delta y_n|+|\Delta y_t|) > M_f(x) - 2\epsilon > 0
\end{equation*}

So for all $\Delta x$ such that $||\Delta x||_p<\epsilon$, and $x'=x+\Delta x$, $x'$ is  classified as $t$.

In \cite{li2019preventing}, authors report a provable robustness of $\sqrt{2}$. However, in their case the global classifier with the $N$ outputs is 1-lipschitz meaning that the $f_i$ have a lipschitz constant lesser than $1$. Then, in their case the maximal value of $M_f(x)$ is lesser than the one of our network, making the comparison of the robustness provable constants not possible directly. 

\section{Experiments : additional results}
\subsection{Networks architecture}
In order to have a fair comparison of the competitors, we use the same architectures for the neural network given a dataset. The architectures are described in Tables \ref{table:MNIST_dense_arch}, \ref{table:MNIST_CNN_arch}, \ref{table:CIFAR_CNN_arch}  and \ref{table:CelebA_CNN_arch}. The activation functions, pooling functions, and normalization functions are described in Tables \ref{table:arch_features}, the optimizations parameters in Table \ref{table:opti_params} and the attacks parameters in Table \ref{table:attach_param}.
\label{sec:networks_architecture}

\begin{table}[]
\centering
\begin{tabular}{|l|l|l|l|}
\hline
Layer   & Number of neurons & Kernel & Output Size \\ \hline
Input   &       N/A            &   N/A     &   784x1          \\ \hline
dense   &       256            &    N/A     &     256        \\ \hline
dense   &       256            &    N/A     &     256        \\ \hline
output   &       10            &    N/A     &     10        \\ \hline
\end{tabular}
\caption{MNIST dense general architecture}
\label{table:MNIST_dense_arch}
\end{table}

\begin{table}[]
\centering
\begin{tabular}{|l|l|l|l|}
\hline
Layer   & Number of neurons & Kernel & Output Size \\ \hline
Input   &       N/A            &   N/A     &   28x28x1          \\ \hline
Conv    &    16               &   3x3     & 28x28x16            \\ \hline
pooling &      N/A             &   2x2     &    14x14x16         \\ \hline
Conv    &      32             &  3x3       &      14x14x32          \\ \hline
pooling &      N/A             &   2x2     &    7x7x32         \\ \hline
dense   &       100            &    N/A     &     100        \\ \hline
output   &       10            &    N/A     &     10        \\ \hline
\end{tabular}
\caption{MNIST CNN general architecture}
   \label{table:MNIST_CNN_arch}
\end{table}

\begin{table}[]
\centering
\begin{tabular}{|l|l|l|l|}
\hline
Layer   & Number of neurons & Kernel & Output Size \\ \hline
Input   &       N/A            &   N/A     &   32x32x3          \\ \hline
Conv    &    128               &   3x3     & 32x32x128         \\ \hline
Conv    &      128             &  3x3       &     32x32x128        \\ \hline
pooling &      N/A             &   2x2     &    16x16x128         \\ \hline
Conv    &    256               &   3x3     & 16x16x256             \\ \hline
Conv    &      256             &  3x3       &     16x16x256          \\ \hline
pooling &      N/A             &   2x2     &    8x8x256         \\ \hline
Conv    &    512               &   3x3     &  8x8x512           \\ \hline
Conv    &      512             &  3x3       &       8x8x512           \\ \hline
pooling &      N/A             &   2x2     &    4x4x512          \\ \hline
dense   &       512            &    N/A     &     512        \\ \hline
dense   &       512            &    N/A     &     512        \\ \hline
output   &       10            &    N/A     &     10        \\ \hline
\end{tabular}
\caption{CIFAR CNN general architecture}
   \label{table:CIFAR_CNN_arch}
   
\end{table}

\begin{table}[]
\centering
\begin{tabular}{|l|l|l|l|}
\hline
Layer   & Number of neurons & Kernel & Output Size \\ \hline
Input   &       N/A            &   N/A     &   128x128x3          \\ \hline
Conv    &    16               &   3x3     & 128x128x16          \\ \hline
Conv    &      16             &  3x3       &     128x128x16         \\ \hline
Conv    &      16             &  3x3       &     128x128x16        \\ \hline
pooling &      N/A             &   2x2     &   64x64x16           \\ \hline
Conv    &    32               &   3x3     & 64x64x32          \\ \hline
Conv    &      32             &  3x3       &    64x64x32        \\ \hline
Conv    &      32             &  3x3       &     64x64x32         \\ \hline
pooling &      N/A             &   2x2     &    16x16x32          \\ \hline
Conv    &    64               &   3x3     & 16x16x64          \\ \hline
Conv    &      64             &  3x3       &    16x16x64      \\ \hline
Conv    &      64             &  3x3       &    16x16x64       \\ \hline
pooling &      N/A             &   2x2     &    8x8x64          \\ \hline
Conv    &    128               &   3x3     & 8x8x128         \\ \hline
Conv    &      128             &  3x3       &     8x8x128         \\ \hline
Conv    &      128             &  3x3       &     8x8x128         \\ \hline
pooling &      N/A             &   2x2     &    4x4x128           \\ \hline
Conv    &    256               &   3x3     & 4x4x256         \\ \hline
Conv    &     256             &  3x3       &     4x4x256         \\ \hline
Conv    &      256             &  3x3       &     4x4x256         \\ \hline
pooling &      N/A             &   2x2     &    2x2x256          \\ \hline
dense   &       512            &    N/A     &     512        \\ \hline
dense   &       512            &    N/A     &     512        \\ \hline
output   &       10            &    N/A     &     10        \\ \hline
\end{tabular}
\caption{CelebA CNN general architecture}
  \label{table:CelebA_CNN_arch}  
\end{table}

\begin{table}[]
\centering
\begin{tabular}{|c|c|c|c|c|c|}
\hline
\textbf{Network} & \textbf{Conv activation} & \textbf{Dense activation} & \textbf{Output activation} & \textbf{Pooling} & \textbf{Orthonormalization} \\ \hline
    $Adv$     &        ReLU       &       ReLU              &       softmax                 &       Maxpooling           &             None                \\ \hline
   $1LIP$ &           ReLU       &        ReLU             &           softmax                 &         Maxpooling         &             Björck        \\ \hline
$GNP_{log}$          &GroupSort2  &    Fullsort                &         softmax                   & 2-norm &      Björck              \\ \hline
$GNP_{hin}$          &GroupSort2  &    Fullsort                &         linear                   & 2-norm &      Björck              \\ \hline
$hKR$          &GroupSort2  &    Fullsort                &         linear                   & 2-norm &      Björck              \\ \hline
\end{tabular}
\caption{Algorithms specific features}
  \label{table:arch_features}  
\end{table}

\begin{table}[]
\centering
\begin{tabular}{|c|c|c|c|c|c|c|}
\hline
\textbf{Dataset} & \textbf{Optimizer} & \textbf{Steps per epoch} & \textbf{Nb epochs} & \textbf{Learning rate} &\textbf{Batch size} &  \textbf{Augmentation} \\ \hline
MNIST dense     &        Adam       &       60000              &       100                 &       0.01     & 256&             no                \\ \hline
MNISTY conv &           Adam       &        60000             &           100                 &         0.01  &  256     &             no        \\ \hline
CIFAR 10          &Adam  &    45000                &         100                   & .00001 & 256 &    no              \\ \hline
CELEB A        &Adam  &    10000                &         200                   & 0.0005 &   64  & yes              \\ \hline
\end{tabular}
\caption{Optimization parameters}
 \label{table:opti_params}  
\end{table}

\begin{table}[]
\centering
\begin{tabular}{|c|c|c|c|c|}
\hline
\textbf{Dataset} & \textbf{$l_2$ deepfool} & \textbf{$l_2$ FGM} & \textbf{$l_2$ PGD} & \textbf{$l_2$ CW} \\ \hline
MNIST dense     &  \vtop{ \hbox{\strut   $\epsilon \in \mathbb{R}^+$}\hbox{\strut 2000 attacks}  }   
&     \vtop{ \hbox{\strut   $\epsilon \text{ from } 0.1 \text{ to } 7.9 \text{ st. } 0.1$}\hbox{\strut 500 attacks}  }             
&       \vtop{ \hbox{\strut  $\epsilon \text{ from } 0.1 \text{ to } 7.9 \text{ st. } 0.1$}\hbox{\strut 500 attacks}  }        
&       \vtop{ \hbox{\strut   $\epsilon \in [0, 1, 2, 4, 6, 8]$}\hbox{\strut 500 attacks}  }              \\ \hline

MNIST conv &  \vtop{ \hbox{\strut   $\epsilon \in \mathbb{R}^+$}\hbox{\strut 2000 attacks}  }   
&     \vtop{ \hbox{\strut   $\epsilon \text{ from } 0.1 \text{ to } 7.9 \text{ st. } 0.1$}\hbox{\strut 500 attacks}  }             
&       \vtop{ \hbox{\strut  $\epsilon \text{ from } 0.1 \text{ to } 7.9 \text{ st. } 0.1$}\hbox{\strut 500 attacks}  }        
&       \vtop{ \hbox{\strut   $\epsilon \in [0, 1, 2, 4, 6, 8]$}\hbox{\strut 500 attacks}  }              \\ \hline

CIFAR 10         &  \vtop{ \hbox{\strut   $\epsilon \in \mathbb{R}^+$}\hbox{\strut 2000 attacks}  }   
&     \vtop{ \hbox{\strut   $\epsilon \text{ from } 0.1 \text{ to } 8 \text{ st. } 0.2$}\hbox{\strut 500 attacks}  }             
&       \vtop{ \hbox{\strut  $\epsilon \text{ from } 0.1 \text{ to } 8 \text{ st. } 0.2$}\hbox{\strut 500 attacks}  }        
&       \vtop{ \hbox{\strut   $\epsilon \in [0, 1, 2, 4, 6, 8]$}\hbox{\strut 500 attacks}  }              \\
\hline

CELEB A       &  \vtop{ \hbox{\strut   $\epsilon \in \mathbb{R}^+$}\hbox{\strut 2000 attacks}  }   
&     \vtop{ \hbox{\strut   $\epsilon \in [2,5,7]$}\hbox{\strut 500 attacks}  }            
&       \vtop{ \hbox{\strut   $\epsilon \in [2,5,7]$}\hbox{\strut 500 attacks}  }        
&       \vtop{ \hbox{\strut   $\epsilon \in [2,5,7]$}\hbox{\strut 500 attacks}  }              \\ \hline
\end{tabular}
\caption{$\epsilon$ values for the different dataset and attacks}
 \label{table:attach_param}  
\end{table}

\end{document}